\PassOptionsToPackage{table,dvipsnames}{xcolor}    
\PassOptionsToPackage{hidelinks}{hyperref}         

\documentclass{article}

\usepackage[utf8]{inputenc}
\usepackage{amsmath,amssymb,mathtools,amsthm}
\usepackage{xcolor}
\usepackage{graphicx}
\usepackage{subfigure}
\usepackage{booktabs}
\usepackage{caption}
\usepackage{placeins}
\usepackage{float}
\FloatBarrier
\usepackage[most]{tcolorbox}
\usepackage{tikz}

\usepackage{natbib}
\usepackage{doi}
\usepackage{hyperref} 
\usepackage[capitalize,noabbrev]{cleveref}

\usepackage[accepted]{icml2025}

\theoremstyle{plain}

\theoremstyle{definition}

\theoremstyle{remark}

\usepackage[textsize=tiny]{todonotes}

\icmltitlerunning{Regional Tiny Stories: Using Small Models to Compare Language Learning and Tokenizer Performance}

\begin{document}


\newcommand{\ColorEvalLossCell}[2]{
  \begingroup
    \def\val{#1}%
    \def\basecolor{#2}%
    \pgfmathsetmacro{\distance}{abs((#1 - 0.2) / (2 - 0.2))} 
    \pgfmathsetmacro{\opacityRaw}{90 - (\distance * 80)} 
    \pgfmathsetmacro{\opacity}{max(0, min(\opacityRaw, 100))}
    \edef\colorcommand{\noexpand\cellcolor{white!\opacity!\basecolor}}%
    \colorcommand\val%
  \endgroup
}

\newcommand{\ColorGeneralCell}[2]{
  \begingroup
    \def\val{#1}%
    \def\basecolor{#2}%
    \pgfmathsetmacro{\distance}{(#1 - 6.5) / (9.5 - 6.5)} 
    \pgfmathsetmacro{\opacityRaw}{25 + (\distance * 75)} 
    \pgfmathsetmacro{\opacity}{max(0, min(\opacityRaw, 100))}
    \edef\colorcommand{\noexpand\cellcolor{white!\opacity!\basecolor}}%
    \colorcommand\hspace{-0.5em}\val%
  \endgroup
}

\newcommand{\ColorModelSizeCell}[2]{
  \begingroup
    \def\val{#1}%
    \def\basecolor{#2}%
    \pgfmathsetmacro{\minval}{0}   
    \pgfmathsetmacro{\maxval}{250} 
    \pgfmathsetmacro{\distance}{(\val - \minval) / (\maxval - \minval)} 
    \pgfmathsetmacro{\opacityRaw}{90 - (\distance * 80)} 
    \pgfmathsetmacro{\opacity}{max(0, min(\opacityRaw, 100))}
    \edef\colorcommand{\noexpand\cellcolor{white!\opacity!\basecolor}}%
    \colorcommand\hspace{-1em}\val%
  \endgroup
}

\twocolumn[
\icmltitle{Regional Tiny Stories: Using Small Models to Compare Language Learning and Tokenizer Performance}



\icmlsetsymbol{equal}{*}
\begin{icmlauthorlist}
\icmlauthor{Nirvan Patil}{equal,group1}
\icmlauthor{Malhar Abhay Inamdar}{equal,group1}
\icmlauthor{Agnivo Gosai}{equal,group1}
\icmlauthor{Guruprasad Pathak}{comp}
\icmlauthor{Anish Joshi}{comp}
\icmlauthor{Aryan Sagavekar}{comp}
\icmlauthor{Anish Joshirao}{comp}
\icmlauthor{Raj Dandekar}{group3}
\icmlauthor{Rajat Dandekar}{group3}
\icmlauthor{Sreedath Panat}{group3}
\end{icmlauthorlist}

\icmlaffiliation{group1}{BITS Goa, Pune Institute of Computer Technology, Independent Researcher, India}
\icmlaffiliation{comp}{PC COE, Pune, India}
\icmlaffiliation{group3}{Vizuara AI Labs, Pune, India}

\icmlcorrespondingauthor{Nirvan Patil}{nirvan.ajit.patil@gmail.com}
\icmlcorrespondingauthor{Malhar Abhay Inamdar}{malhar.inamdar.097@gmail.com}
\icmlcorrespondingauthor{Agnivo Gosai}{agnivo2007@gmail.com}
\icmlcorrespondingauthor{Raj Dandekar}{raj@vizuara.com}

\begin{center}
\end{center}
\icmlkeywords{Machine Learning, ICML}

\vskip 0.3in]



\printAffiliationsAndNotice{\icmlEqualContribution} 

\begin{abstract}
The 2023 TinyStories study developed an English dataset that allows Small Language Models (SLMs) with 1–10 million parameters to produce coherent outputs matching those of LLMs. Our research expands this framework by creating translated as well as synthetically generated datasets in Indian languages. Using this new dataset , we demonstrate that SLMs efficiently process regional languages with significantly fewer parameters than LLMs, and additionally offer a  complementary framework for ``inference-based evaluation" of tokenization strategies and linguistic complexity. Our analysis reveals that language-specific tokenizers outperform general-purpose ones for Indian languages. Empirical validations, supported by information-theoretic and morphological analyses, provide insights into the superior performance of Hindi models over Marathi and Bengali. The study uncovers distinct cross-linguistic patterns: Bengali emphasizes creativity, Hindi excels in context understanding and grammar with model scaling, and Marathi requires larger models to capture its unique linguistic features. Optimal parameter allocation varies, with Hindi benefiting more from wider architectures and Bengali favoring a balanced approach. We also show that quality synthetic datasets outperform translated content for training SLMs by 15-30 \% . These findings advance both the practical application of SLMs to underserved languages and our theoretical understanding of neural language development. 
\end{abstract}

\section{Introduction}

Recent advances in Large Language Models (LLMs) have predominantly focused on scaling architectures to multi-billion parameters \cite{brown2020languagemodelsfewshotlearners,chowdhery2022palmscalinglanguagemodeling}, driven by the generally accepted notion that increased model size directly correlates with improved performance, which relies on ever-increasing compute and data requirements \cite{hoffmann2022trainingcomputeoptimallargelanguage}. However, \citeauthor{eldan2023tinystoriessmalllanguagemodels} (\citeyear{EnglishTinyStories}) challenged this paradigm through their TinyStories framework, demonstrating that Small Language Models (SLMs) with fewer than 50M parameters can achieve noteworthy performance when trained on carefully constructed but much smaller datasets. This is also observed in children who are generally exposed to no more than 100 million words by the age of 13 \cite{Gilkerson2017MappingTE}, showing remarkable learning efficiency in comparison to leading LMs. By generating synthetic stories using preschool-level vocabulary through GPT-3.5 and GPT-4, the Tinystories paper established three fundamental findings: (1) coherent text generation and basic reasoning capabilities can emerge in significantly smaller architectures than previously theorized, (2) language capabilities develop hierarchically, beginning with grammatical structure and progressing through contextual consistency to creative generation, and (3) architectural choices significantly impact specific competencies, with model width correlating to knowledge retention and depth to contextual understanding. These results suggest that the field's focus on massive architectures may be unnecessary for many language modeling tasks, opening new possibilities for efficient, targeted model development. 
\newline

In last two years, SLMs and modeling low-resource languages have gained traction as seen in the BabyLM challenge, proposed by \citeauthor{warstadt2023papersbabylmchallenge} (\citeyear{warstadt2023papersbabylmchallenge}), which encourages participants to focus on cognitive modeling and effective language model pre-training, keeping in mind data constraints that mirror human development. Consequently,  \citeauthor{muckatira2024emergentabilitiesreducedscalegenerative} (\citeyear{muckatira2024emergentabilitiesreducedscalegenerative}), found that smaller models trained on simplified vocabulary outperformed larger models trained on complete datasets at zero-shot tasks, indicating that data complexity significantly influences zero-shot capabilities in smaller models.  Interestingly, \citeauthor{boughorbel2024improvinglanguagemodelstrained} (\citeyear{boughorbel2024improvinglanguagemodelstrained}), reported Tinystories inspired Arabic SLMs, where initial models trained on  translated data exhibited various quality and task-specific issues, whereas further pre-training with a small amount (1\%) of high-quality synthetic Arabic stories generated by GPT-4 significantly improved performance. In another recent study, \citeauthor{theodoropoulos2024berttimestoriesinvestigatingrole}, (\citeyear{theodoropoulos2024berttimestoriesinvestigatingrole}), show that mixing high quality synthetic data with a subset of Tinystories, had modest or no improvements in the output of an LTG-BERT \cite{samuel-etal-2023-trained} model when compared to outputs of GPT-Neo \cite{Black2021GPTNeoLS} which was trained on the original Tinystories only. However, the overall performance of these LMs were capped around ~$\approx 70\%$, indicating a huge scope of improvement as well as illuminating the roles played by data type and quality.

While TinyStories presents compelling evidence for English language modeling with small architectures, two critical questions remain unexplored. 
\begin{itemize}
    \item First, can this modeling paradigm be extended effectively to Indian languages and is the quality of tokenizer the determining factor behind high-quality output? The development of current language models shows a significant bias towards English \cite{wang2024comprehensivesurveysmalllanguage}, comprising 30-60\% of training data in most large-scale models. The often used BLiMP framework \cite{warstadt2023blimpbenchmarklinguisticminimal} applies to linguistic knowledge evaluation for grammatical phenomenon in English.
    \item Second, can we leverage the TinyStories framework as a comparative tool to analyze the inherent complexities across different languages? The hypothesis being that the minimum parameter count required for effective modeling might serve as a proxy for language complexity.
\end{itemize}

To address these questions, we focus on three major Indian languages with diverse linguistic characteristics: Hindi (spoken by approximately 600-700 million people), Marathi (83-85 million speakers), and Bengali (97-100 million speakers). These languages present an ideal test case due to their significant speaker populations and distinct linguistic features. Despite their importance, there exists limited comparative analysis of their inherent complexities, particularly in the context of neural language modeling. Our work makes several key contributions :
\begin{itemize}
    \item We demonstrate the successful adaptation of the TinyStories SLM paradigm to these three Indian languages, detailing effective pre-training steps and demonstrating substantial inference quality with significantly smaller model sizes than current state-of-the-art approaches. 
    \item We establish a novel methodology for comparing linguistic complexity across languages using the TinyStories evaluation framework. 
    \item We provide comprehensive analysis of tokenization efficiency across these languages, comparing standard approaches from OpenAI with specialized Indian language tokenizers like Sarvam and SUTRA. 
    \item Consequently we provide an alternative framework for evaluating tokenizers for specific language use cases based on SLM inference quality, compared to established benchmarks like tokens/word.
    \item  Our training data analysis shows substantial lexical differences (low BLEU $\approx$ 0.078) thus confirming variety essential for training, despite perceived semantic equivalence (high BERTScore $\approx$ 1.0) , highlighting the limitation of standard metrics (ROUGE = 0 ) which works for English but fails for morphologically richer Indian languages. 
    \item We show that synthetic dataset generation outperforms simple translation-based approaches, with regards to inference quality for these languages.
    \item Finally, we report and release our training data of  $\approx$ 10M synthetic and translated stories in three Indian langauges along with the trained models for the broader community.
\end{itemize}

These findings have significant implications for both theoretical linguistics and practical applications in low-resource language modeling. Our results suggest that effective language modeling for Indian languages may not require the massive architectures currently considered standard but need task-specific quality dataset, potentially democratizing access to language technology for underrepresented languages. 

\section{Methodology: Data generation, training and evaluation experiments}
\subsection{Training data preparation}

Our research extends the TinyStories framework \cite{EnglishTinyStories} to explore simple, constrained narratives in multiple Indian languages through a two-phase approach: translating the original English TinyStories dataset \cite{eldan2023tinystoriessmalllanguagemodels} into Indian languages, followed by generating additional synthetic data using LLMs while maintaining the original methodology's constraints (Fig. 1).

\begin{figure}[h]
    \centering
    \includegraphics[width=0.5\textwidth]{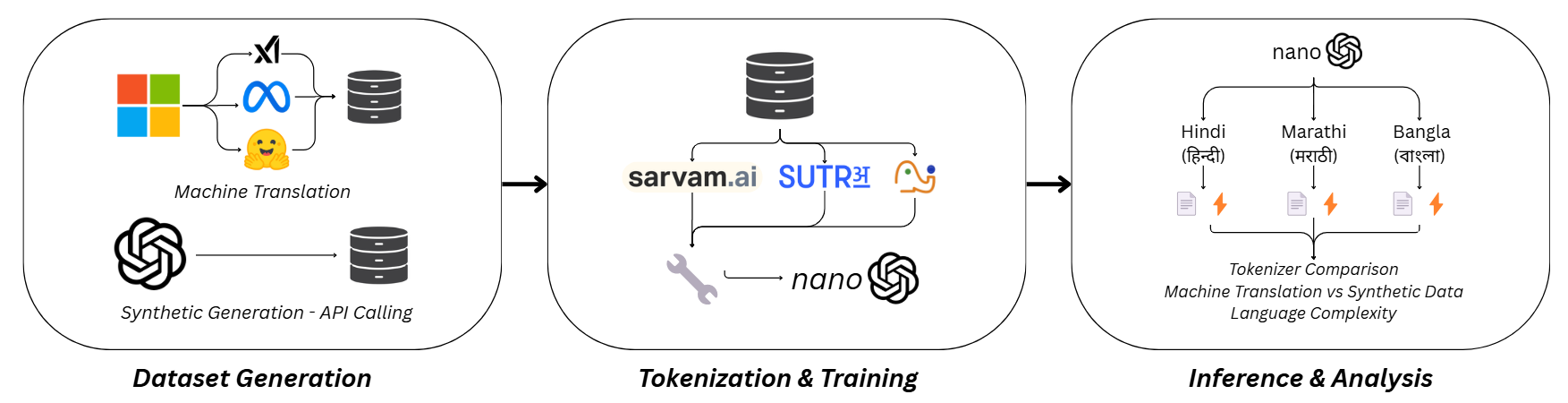}
    \caption{Schematic of model pipeline. (A) Dataset prepared through machine translation as well as generation using LLM, (B) Indic tokenizers are used to preprocess the Indian language stories, (C) A decoder only transformer architecture (nanoGPT) is used to train the model in each language ; Inference is evaluated by LLM on linguistic attributes}
    \label{Fig. 1}
\end{figure}
\subsubsection{Translated data}
Previously \citeauthor{doshi-etal-2024-pretraining} \citeyearpar{doshi-etal-2024-pretraining}, demonstrated that machine-translated filtered data can be used to train language models for Indian languages, which can match the performance of models trained on clean native data. Likewise NLLB-3B MT has been used to translate Tiny Stories into Arabic \cite{boughorbel2024improvinglanguagemodelstrained}. Hence, using translated data for training language models is not uncommon. For dataset creation, we first translated the complete TinyStories dataset of approximately 2.0 million short stories from English to Hindi and Bengali using a combination of NLLB-200-3B and Google Translate. We verified a random subset of 1,000 stories per language using LLM to assess semantic preservation, grammatical accuracy, cultural appropriateness, and consistency in reading level.
\subsubsection{Synthetic data}
We generated synthetic data by prompting GPT-4o-mini. The process began with vocabulary generation using GPT-4, creating word lists exceeding 700 entries each for nouns, verbs, and adjectives in the three languages. After filtering through GPT-3.5 to remove inappropriate content, we retained approximately 300 carefully curated words in each category. For narrative control, we generated generic features instead of specific story elements to reduce repetition patterns.

The prompt generation feature incorporated a unique identification system that combined linguistic elements systematically and prevented duplicates, successfully eliminating about 37,500 potential duplicate prompts from the 3M target dataset. After evaluating multiple models including GPT-4, LLaMA-3.1 70B, and Claude 3.5 Sonnet, we selected GPT-4o-mini based on its optimal balance of quality and generation efficiency, achieving an 8.5/10 average score based on story completeness, grammar, fluency, creativity, using GPT-4o as judge.

Quality assessment was performed using GPT-4 as the evaluation model, examining completeness, grammar, fluency, and creativity. The final implementation used complexity level 2+ prompts with expanded word limits, consistently producing the highest quality stories while maintaining generation efficiency. This approach yielded impressive evaluation scores averaging 8.73 across all metrics (details in Appendix D).

The final dataset includes 1.8M translated stories each in Hindi and Bengali, along with 2.2M new synthetic stories for Hindi, Bengali and Marathi. All content was standardized in JSON format with comprehensive metadata. 

\subsection{Training data evaluation}
Our analysis of training data revealed significant limitations in traditional evaluation metrics when applied to Indian languages. As documented in Appendix E, ROUGE scores \cite{lin-2004-rouge} consistently registered zero for semantically similar Bengali stories, highlighting a fundamental challenge in evaluating non-English text generation. Multiple metrics provided complementary insights: BERTScore \cite{zhang2020bertscoreevaluatingtextgeneration} values near 1.0 confirmed strong semantic equivalence between examples, while BLEU scores \cite{10.3115/1073083.1073135} remained consistently low (averaging 0.078). METEOR \cite{banerjee-lavie-2005-meteor} offered middle-ground assessment (averaging 0.153) by recognizing synonyms and word variations.

This pattern—high semantic similarity (BERTScore) with low lexical overlap (BLEU/METEOR)—indicates our dataset contains diverse lexical expressions of similar concepts. Such divergence occurs because morphologically rich Indian languages permit extensive variation in expressing equivalent meanings. The apparent metric ``anomaly" actually reveals a strength in our dataset: stories maintain semantic coherence while exhibiting rich linguistic variation, precisely the characteristics needed for robust language modeling. Rather than memorizing phrases, models learn to understand concepts expressed through diverse vocabulary and structures. This finding underscores both the challenge of evaluating Indian language generation and the benefit of our approach, which produces semantically coherent yet lexically diverse training examples that foster more generalizable language understanding.

\subsection{Tokenizer, model and inference evaluation}
The tokenizers were chosen specifically for Indian language modeling, for e.g. Sarvam \cite{SARVAM1} and SUTRA by TWO AI \cite{bendale2024SUTRAscalablemultilinguallanguage} which incorporate tokens for language-specific elements and formatting, and results were compared with OpenAI's Tiktoken \cite{Tiktoken-OpenAI}. For e.g., Sarvam-1's advanced tokenizer achieves near-English token fertility rates (1.4-2.1 tokens per word) for Indic scripts, significantly improving efficiency and performance compared to traditional multilingual LLMs that struggle with high token fertility in Indian languages. Modern language model tokenizers differ in vocabulary size: OpenAI's Tiktoken (GPT2) uses 50,257 tokens, SUTRA has about 256,000 tokens, and Sarvam features 68,096 tokens with 4,096 reserved for future use. 

We built on TinyStories using modified nanoGPT code \cite{Karpathy-2022}, implementing decoder-only transformers with 8 attention heads at various parameter sizes. All models trained for 5001 epochs with 2.5\% of data reserved for testing. The inference evaluation, across chosen linguistic attributes, was conducted by GPT-4o, following the previously established LLM as a judge framework \cite{EnglishTinyStories,boughorbel2024improvinglanguagemodelstrained}.

\section{Results}
\subsection{Insights from our evaluation method}

\textbf{Model Architecture and Scaling Dynamics}

Results for Hindi, Marathi, and Bengali models, trained using Sarvam tokenizer and synthetic data (Tables 1-3), reveal both systematic model scaling patterns and significant language-specific characteristics. The analysis demonstrates that increasing model size from $\approx$ 4.5M to 153M parameters yields consistent performance improvements across all three languages, with the most substantial gains occurring in the range of $\approx$ 5M to 73M parameters.

Our efficiency-performance analysis identifies an optimal configuration of 512 hidden units and 6 layers, totaling around 54M parameters. This architecture delivers strong performance across Hindi (8.158), Bengali (8.016), and Marathi (7.807) while keeping computational demands manageable. The relationship between layer depth and performance consistently shows optimal results at moderate depths, particularly with 6 layers across all languages.

Basic linguistic capabilities can be achieved with smaller models (4.46M-10M parameters), whereas complex story generation requires significantly larger models (41M+ parameters). This trend is consistent across the three languages, although Marathi typically needs 20-30\% more parameters to match the performance levels of Hindi and Bengali. Beyond the optimal configuration, performance gains diminish relative to the quadratic growth in parameter count, highlighting important considerations for model design and deployment.

\textbf{Cross-lingual Performance and Metric Analysis}

Model evaluation loss consistently decreases across all languages, with Hindi dropping from about 1.4 to 0.5, indicating improved optimization in larger models. This trend is most pronounced in Hindi and Bengali, suggesting better learning. Architectural scaling impacts Hindi, Bengali, and Marathi differently. Hindi models achieve the best performance (8.164 at 1024/7 configuration) with notable improvements in grammar (8.910) and fluency (8.580), and steady gains in overall score as model size increases. Bengali models show similar scaling characteristics (8.037 at 1024/7), hinting at structural similarities between these languages and indeed they belong to fusional languages. Conversely, Marathi which is an agglutinative language, requires still larger models for similar performance,  indicating its linguistic features need more model capacity for effective processing. 

Major observations: 
\begin{itemize}
    \item Context understanding capabilities vary significantly: Hindi and Bengali demonstrate robust improvements with scale (reaching approximately 7.7 at 1024/7), while Marathi exhibits more modest gains, pointing to distinct challenges in capturing contextual relationships.
\end{itemize}
\begin{itemize}
    \item Grammatical competence shows the most dramatic scaling improvements across all languages, though from different baselines – Hindi and Bengali achieve strong performance (approximately 8.4 at 64/6) even at modest scales, while Marathi requires larger models (512/2) to reach comparable accuracy.
\end{itemize}
\begin{itemize}
    \item Optimal parameter allocation differs by language: Hindi models keep gaining with wider architecture (more embeddings), while Bengali models prefer a balanced approach with moderate width and depth.
    \item Interestingly, increasing embedding dimensions beyond 512, largely stymies the percentage improvement in attributes like context, grammar and overall score. This is most evident in Marathi and Bengali but further increase in layer depth allows some improvements.
\end{itemize}

Based on 3000 stories per model inference, evaluation metrics show both universal patterns (strong correlations between creativity-quality, grammar-quality, and completeness-fluency) and language-specific relationships (Bengali emphasizes creativity, Hindi shows weaker context-completeness links, and Marathi uniquely correlates context with grammar). More details are provided in Appendix B.2-B.4.

\textbf{Emergence of Linguistic Capabilities}

The development of critical capabilities follows a consistent pattern across languages, revealing a clear link between model capacity and linguistic competence. Basic grammatical competence emerges in models with 64-128 hidden units (4.46M-10M parameters), achieving grammar scores of about 8.4 for Hindi and Bengali. 
\begin{table*}[ht!]
\footnotesize
\centering
\begin{tabular}{c c c c c c c c c c}
\toprule
\textbf{Hidden Size} & \textbf{Layer} & \textbf{Model Size} & \textbf{Eval Loss} & \textbf{Context} & \textbf{Completeness} & \textbf{Creativity} & \textbf{Fluency} & \textbf{Grammar} & \textbf{Overall} \\
\midrule
\textbf{64} & \textbf{2}   & \ColorModelSizeCell{4.46}{red}  & \ColorEvalLossCell{1.408}{red} & \ColorGeneralCell{5.665}{red} & \ColorGeneralCell{6.826}{red} & \ColorGeneralCell{7.217}{red} & \ColorGeneralCell{7.472}{red} & \ColorGeneralCell{7.969}{red} & \ColorGeneralCell{7.030}{red} \\
\textbf{64} & \textbf{6}   & \ColorModelSizeCell{4.65}{red}  & \ColorEvalLossCell{1.182}{red} & \ColorGeneralCell{6.412}{red} & \ColorGeneralCell{7.122}{red} & \ColorGeneralCell{7.314}{red} & \ColorGeneralCell{7.901}{red} & \ColorGeneralCell{8.446}{red} & \ColorGeneralCell{7.439}{red} \\
\textbf{64} & \textbf{12}  & \ColorModelSizeCell{5.00}{red}  & \ColorEvalLossCell{1.057}{red} & \ColorGeneralCell{6.374}{red} & \ColorGeneralCell{7.227}{red} & \ColorGeneralCell{7.390}{red} & \ColorGeneralCell{7.959}{red} & \ColorGeneralCell{8.450}{red} & \ColorGeneralCell{7.480}{red} \\
\midrule
\textbf{512} & \textbf{2}  & \ColorModelSizeCell{41.00}{red} & \ColorEvalLossCell{0.654}{red} & \ColorGeneralCell{7.054}{red} & \ColorGeneralCell{7.661}{red} & \ColorGeneralCell{7.705}{red} & \ColorGeneralCell{8.427}{red} & \ColorGeneralCell{8.746}{red} & \ColorGeneralCell{7.919}{red} \\
\textbf{512} & \textbf{6}  & \ColorModelSizeCell{54.00}{red}& \ColorEvalLossCell{0.518}{red} & \ColorGeneralCell{7.734}{red} & \ColorGeneralCell{7.783}{red} & \ColorGeneralCell{7.806}{red} & \ColorGeneralCell{8.554}{red} & \ColorGeneralCell{8.912}{red} & \ColorGeneralCell{8.158}{red} \\
\textbf{512} & \textbf{12} & \ColorModelSizeCell{73.00}{red} & \ColorEvalLossCell{0.519}{red} & \ColorGeneralCell{7.572}{red} & \ColorGeneralCell{7.659}{red} & \ColorGeneralCell{7.718}{red} & \ColorGeneralCell{8.458}{red} & \ColorGeneralCell{8.862}{red} & \ColorGeneralCell{8.054}{red} \\
\midrule
\textbf{1024} & \textbf{2} & \ColorModelSizeCell{94.00}{red} & \ColorEvalLossCell{0.581}{red} & \ColorGeneralCell{7.344}{red} & \ColorGeneralCell{7.798}{red} & \ColorGeneralCell{7.829}{red} & \ColorGeneralCell{8.516}{red} & \ColorGeneralCell{8.825}{red} & \ColorGeneralCell{8.062}{red} \\
\textbf{1024} & \textbf{7} & \ColorModelSizeCell{153.00}{red}& \ColorEvalLossCell{0.513}{red} & \ColorGeneralCell{7.695}{red} & \ColorGeneralCell{7.806}{red} & \ColorGeneralCell{7.830}{red} & \ColorGeneralCell{8.580}{red} & \ColorGeneralCell{8.910}{red} & \ColorGeneralCell{8.164}{red} \\
\bottomrule
\end{tabular}
\caption{\small{\textbf{Hindi} - This table illustrates the hyperparameter configurations and evaluation results for Hindi Stories.  The color coding is such that, the lighter the color, the better the performance. No. of attention heads = 8, tokenizer vocab size = 68096 (Sarvam). Mean scores across 3000 samples are reported for each model configuration. }}
\label{tab:hindi-eval}
\captionsetup{justification=centering}
\vspace{0.3em}

\end{table*}

\definecolor{teal}{rgb}{0.0, 0.5, 0.5}

\begin{table*}[ht!]
\footnotesize
\centering
\begin{tabular}{c c c c c c c c c c}
\toprule
\textbf{Hidden Size} & \textbf{Layer} & \textbf{Model Size} & \textbf{Eval Loss} & \textbf{Context} & \textbf{Completeness} & \textbf{Creativity} & \textbf{Fluency} & \textbf{Grammar} & \textbf{Overall} \\
\midrule
\textbf{64} & \textbf{2}   & \ColorModelSizeCell{4.46}{teal}  & \ColorEvalLossCell{3.7298}{teal} & \ColorGeneralCell{5.618}{teal} & \ColorGeneralCell{6.615}{teal} & \ColorGeneralCell{7.525}{teal} & \ColorGeneralCell{6.823}{teal} & \ColorGeneralCell{7.411}{teal} & \ColorGeneralCell{6.799}{teal} \\
\textbf{64} & \textbf{6}   & \ColorModelSizeCell{4.65}{teal}  & \ColorEvalLossCell{2.843}{teal} & \ColorGeneralCell{6.171}{teal} & \ColorGeneralCell{6.974}{teal} & \ColorGeneralCell{7.435}{teal} & \ColorGeneralCell{7.390}{teal} & \ColorGeneralCell{8.103}{teal} & \ColorGeneralCell{7.215}{teal} \\
\textbf{64} & \textbf{12}  & \ColorModelSizeCell{5.00}{teal}  & \ColorEvalLossCell{2.6244}{teal} & \ColorGeneralCell{6.249}{teal} & \ColorGeneralCell{7.009}{teal} & \ColorGeneralCell{7.288}{teal} & \ColorGeneralCell{7.471}{teal} & \ColorGeneralCell{8.184}{teal} & \ColorGeneralCell{7.240}{teal} \\
\midrule
\textbf{512} & \textbf{2}  & \ColorModelSizeCell{41.00}{teal} & \ColorEvalLossCell{2.3330}{teal} & \ColorGeneralCell{6.934}{teal} & \ColorGeneralCell{7.396}{teal} & \ColorGeneralCell{7.521}{teal} & \ColorGeneralCell{8.002}{teal} & \ColorGeneralCell{8.603}{teal} & \ColorGeneralCell{7.691}{teal} \\
\textbf{512} & \textbf{6}  & \ColorModelSizeCell{54.00}{teal}& \ColorEvalLossCell{2.0761}{teal} & \ColorGeneralCell{7.245}{teal} & \ColorGeneralCell{7.407}{teal} & \ColorGeneralCell{7.553}{teal} & \ColorGeneralCell{8.106}{teal} & \ColorGeneralCell{8.723}{teal} & \ColorGeneralCell{7.807}{teal} \\
\textbf{512} & \textbf{12} & \ColorModelSizeCell{73.00}{teal} & \ColorEvalLossCell{1.8117}{teal} & \ColorGeneralCell{7.281}{teal} & \ColorGeneralCell{7.565}{teal} & \ColorGeneralCell{7.664}{teal} & \ColorGeneralCell{8.156}{teal} & \ColorGeneralCell{8.739}{teal} & \ColorGeneralCell{7.881}{teal} \\
\midrule
\textbf{1024} & \textbf{2} & \ColorModelSizeCell{94.00}{teal} & \ColorEvalLossCell{0.680}{teal} & \ColorGeneralCell{6.728}{teal} & \ColorGeneralCell{7.184}{teal} & \ColorGeneralCell{7.484}{teal} & \ColorGeneralCell{7.687}{teal} & \ColorGeneralCell{8.295}{teal} & \ColorGeneralCell{7.476}{teal} \\
\textbf{1024} & \textbf{7} & \ColorModelSizeCell{153.00}{teal}& \ColorEvalLossCell{0.619}{teal} & \ColorGeneralCell{7.275}{teal} & \ColorGeneralCell{7.152}{teal} & \ColorGeneralCell{7.540}{teal} & \ColorGeneralCell{7.896}{teal} & \ColorGeneralCell{8.625}{teal} & \ColorGeneralCell{7.698}{teal} \\
\bottomrule
\end{tabular}
\caption{\small{\textbf{Marathi} - This table illustrates the hyperparameter configurations and evaluation results for Marathi Stories.  The color coding is such that, the lighter the color, the better the performance. No. of attention heads = 8, tokenizer vocab size = 68096 (Sarvam). Mean scores for 3000 samples are reported for each model configuration. }}
\label{tab:marathi-eval}

\captionsetup{justification=centering}
\vspace{0.3em}

\end{table*}


\definecolor{beige}{rgb}{0.70, 0.34, 0.28}

\begin{table*}[ht]
\footnotesize
\centering
\begin{tabular}{c c c c c c c c c c}
\toprule
\textbf{Hidden Size} & \textbf{Layer} & \textbf{Model Size} & \textbf{Eval Loss} & \textbf{Context} & \textbf{Completeness} & \textbf{Creativity} & \textbf{Fluency} & \textbf{Grammar} & \textbf{Overall} \\
\midrule
\textbf{64} & \textbf{2}   & \ColorModelSizeCell{4.46}{beige}  & \ColorEvalLossCell{1.514}{beige} & \ColorGeneralCell{6.663}{beige} & \ColorGeneralCell{7.097}{beige} & \ColorGeneralCell{7.469}{beige} & \ColorGeneralCell{7.797}{beige} & \ColorGeneralCell{8.424}{beige} & \ColorGeneralCell{7.490}{beige} \\
\textbf{64} & \textbf{6}   & \ColorModelSizeCell{4.65}{beige}  & \ColorEvalLossCell{1.245}{beige} & \ColorGeneralCell{6.533}{beige} & \ColorGeneralCell{7.225}{beige} & \ColorGeneralCell{7.482}{beige} & \ColorGeneralCell{7.975}{beige} & \ColorGeneralCell{8.454}{beige} & \ColorGeneralCell{7.534}{beige} \\
\textbf{64} & \textbf{12}  & \ColorModelSizeCell{5.00}{beige}  & \ColorEvalLossCell{1.136}{beige} & \ColorGeneralCell{6.760}{beige} & \ColorGeneralCell{7.289}{beige} & \ColorGeneralCell{7.563}{beige} & \ColorGeneralCell{7.968}{beige} & \ColorGeneralCell{8.507}{beige} & \ColorGeneralCell{7.617}{beige} \\
\midrule
\textbf{512} & \textbf{2}  & \ColorModelSizeCell{41.00}{beige} & \ColorEvalLossCell{0.693}{beige} & \ColorGeneralCell{7.373}{beige} & \ColorGeneralCell{7.494}{beige} & \ColorGeneralCell{7.644}{beige} & \ColorGeneralCell{8.314}{beige} & \ColorGeneralCell{8.782}{beige} & \ColorGeneralCell{7.922}{beige} \\
\textbf{512} & \textbf{6}  & \ColorModelSizeCell{54.00}{beige} & \ColorEvalLossCell{0.569}{beige} & \ColorGeneralCell{7.507}{beige} & \ColorGeneralCell{7.645}{beige} & \ColorGeneralCell{7.693}{beige} & \ColorGeneralCell{8.420}{beige} & \ColorGeneralCell{8.816}{beige} & \ColorGeneralCell{8.016}{beige} \\
\textbf{512} & \textbf{12} & \ColorModelSizeCell{73.00}{beige} & \ColorEvalLossCell{0.544}{beige} & \ColorGeneralCell{7.525}{beige} & \ColorGeneralCell{7.718}{beige} & \ColorGeneralCell{7.743}{beige} & \ColorGeneralCell{8.450}{beige} & \ColorGeneralCell{8.836}{beige} & \ColorGeneralCell{8.054}{beige} \\
\midrule
\textbf{1024} & \textbf{2} & \ColorModelSizeCell{95.00}{beige} & \ColorEvalLossCell{0.609}{beige} & \ColorGeneralCell{7.407}{beige} & \ColorGeneralCell{7.470}{beige} & \ColorGeneralCell{7.626}{beige} & \ColorGeneralCell{8.293}{beige} & \ColorGeneralCell{8.786}{beige} & \ColorGeneralCell{7.916}{beige} \\
\textbf{1024} & \textbf{7} & \ColorModelSizeCell{157.00}{beige}& \ColorEvalLossCell{0.557}{beige} & \ColorGeneralCell{7.567}{beige} & \ColorGeneralCell{7.639}{beige} & \ColorGeneralCell{7.740}{beige} & \ColorGeneralCell{8.409}{beige} & \ColorGeneralCell{8.832}{beige} & \ColorGeneralCell{8.037}{beige} \\
\bottomrule
\end{tabular}
\caption{\small{\textbf{Bengali} - This table illustrates the hyperparameter configurations and evaluation results for Bengali Stories.  The color coding is such that, the lighter the color, the better the performance. No. of attention heads = 8, tokenizer vocab size = 68096 (Sarvam). Mean scores for 3000 samples are reported for each model configuration.}}
\label{tab:updated-bengali-eval}

\captionsetup{justification=centering}
\vspace{0.3em}

\end{table*}
At 256 hidden units (19M-27M parameters), story completion consistency improves significantly, with scores rising from 6.8 to 7.6 across languages. More advanced capabilities, such as context understanding and creativity, rely more heavily on model capacity. These begin to emerge meaningfully at 512 hidden units (41M-73M parameters), with context scores exceeding 7.5 and creativity metrics nearing 7.8. This threshold is crucial for Marathi, showing marked improvements in contextual processing only at this scale. Larger models (768-1024 hidden units, 85M-153M parameters) continue to enhance creative expression and contextual coherence, albeit with diminishing returns. 

The hierarchical emergence of capabilities suggests that neural story generation follows a structured developmental pattern, similar to language acquisition, where simpler grammatical competencies precede more complex narrative abilities. This view aligns with the idea that language is acquired primarily through social interactions and pattern recognition in speech, rather than being driven by an innate, specialized ``language module." Language learning emerges from general cognitive skills, such as intention-reading and pattern-finding, which underlie both grammatical and narrative development \cite{0b455472-151c-3430-8dee-e4823ceceb8c}.  

\subsection{Comparing inference results of our Regional TinyStories with LLMs}

Comparison with reference large language models (LLMs) provides valuable insights into the current capabilities and limitations of our approach. GPT-4 and SUTRA variants establish strong baselines across all languages, consistently achieving scores between 9.0 and 9.5. While our scaled models approach these performance levels in specific metrics, particularly grammar and fluency, a notable gap persists in context understanding and creativity. SUTRA-Pro demonstrates superior performance compared to SUTRA-Light across all languages, with marked advantages in context understanding (9.60 versus 9.20), suggesting that its architectural improvements benefit contextual processing independently of the target language. 

In Figs. 2 \&\ 3, we show the stories generated from our TinyStories 5M model and that of GPT-4o for Hindi language for the same prompt. The prompt which we gave in Hindi effectively translates to: 
\begin{figure}[H]
    \centering
    \begin{tcolorbox}[colback=gray!15, colframe=gray!25, boxrule=0.5pt, arc=2mm, width=\linewidth]
        ‘‘Once upon a time, there was a small boy. His toy...’’
    \end{tcolorbox}
\end{figure}
\vspace{-0.2cm}
\begin{figure}[!htb]
    \centering
    \includegraphics[width=0.5\textwidth]{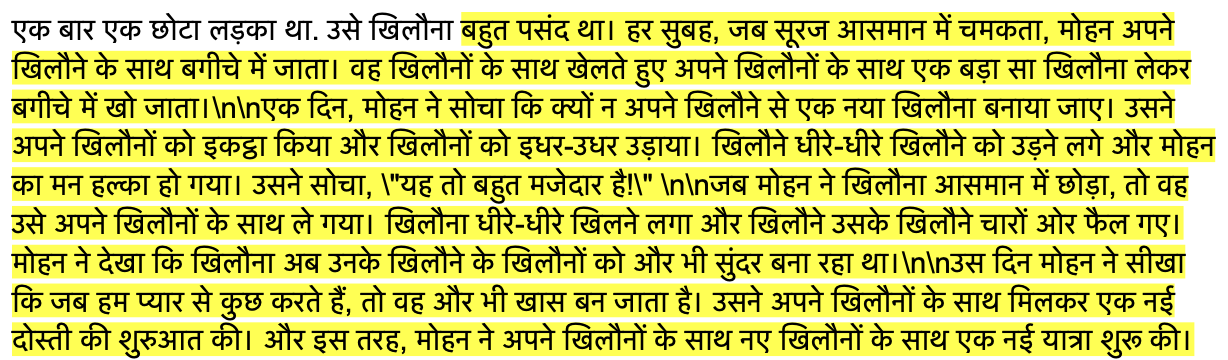}
    \vspace{-0.6cm}
    \caption{Regional TinyStories 5M Hindi model generated story, story after prompt is highlighted}
    \label{fig:Hindi 5M story}
\end{figure}
\vspace{-0.6cm}
\begin{figure}[!htb]
    \centering
    \includegraphics[width=0.5\textwidth]{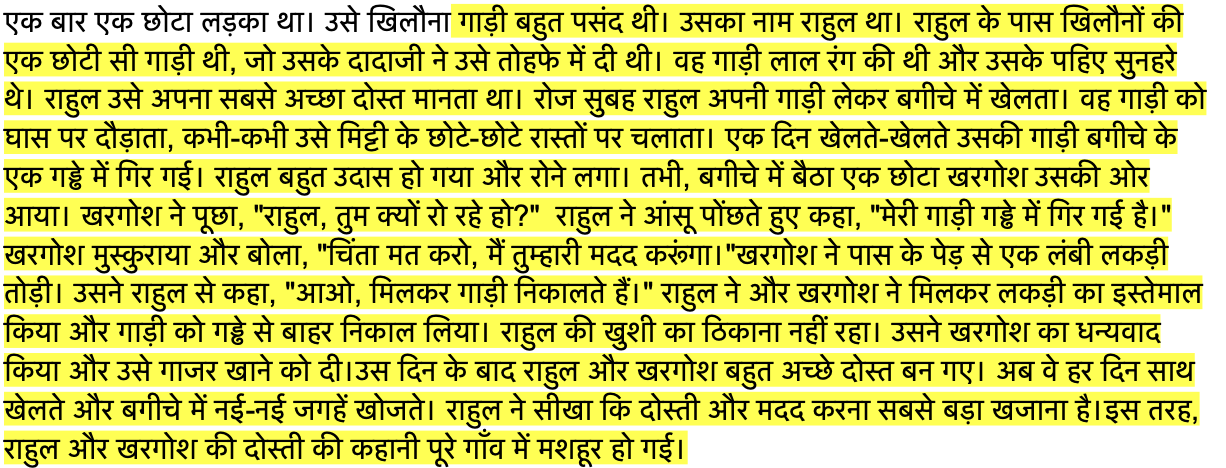}
    \vspace{-0.6cm}
    \caption{GPT4o generated Hindi story}
    \label{fig:Hindi GPT4o story}
\end{figure}

We then ask GPT-4o to qualitatively compare between the 2 stories. Here is the output : 
\begin{tcolorbox}[colback=gray!20, colframe=gray!50, boxrule=0.5pt, arc=2mm]
Both stories follow a similar narrative structure, focusing on a young boy who loves playing with toys. Both stories conclude with a positive resolution, emphasizing companionship and the joy of forming new relationships. Both are written in simple, accessible Hindi, making them suitable for children.
\end{tcolorbox}
This demonstrates that despite having a parameter count nearly 1,000,000 times lower than GPT-4o, we successfully generate coherent and fluent stories with clear messages. Appendix C includes  SLM vs LLM generated Marathi and Bengali story results.

\subsection{Regional tokenizers perform better}

Table 4 compares three tokenizers—Sarvam, SUTRA, and Tiktoken—across Hindi, Marathi, and Bengali using both quantitative and qualitative metrics based on stories generated by our 54M parameter models. Our analysis reveals a striking pattern: Tiktoken consistently achieves the lowest evaluation loss across all languages (Hindi: 0.149, Marathi: 0.167, Bengali: 0.135), suggesting superior perplexity minimization. However, this advantage doesn't translate to generation quality, where Tiktoken underperforms on all subjective dimensions. Indian language-specific tokenizers demonstrate superior performance in generation quality. Sarvam achieves the highest overall scores for all languages (Hindi: 8.158, Marathi: 7.807, Bengali: 8.016), particularly excelling in context understanding and narrative completeness. SUTRA follows closely, with strengths in grammatical accuracy.

The performance gap is most pronounced in context awareness (+0.56 points average for Sarvam over Tiktoken) and fluency (+0.63 points average). This suggests regionally specialized tokenizers better capture semantic cohesion, idiomatic expressions, and structural nuances. These findings align with research showing general-purpose tokenizers introduce significant biases in non-English languages, requiring up to 15 times more tokens for equivalent content \cite{petrov2023languagemodeltokenizersintroduce}. The superior performance of language-specific tokenizers can be attributed to several factors: (1) more efficient subword segmentation aligned with morphological boundaries, (2) better handling of script-specific features in Devanagari and Bengali scripts, and (3) vocabulary coverage optimized for the linguistic distributions of these languages. These advantages are particularly evident in the grammar scores, where both Sarvam and SUTRA demonstrate robust handling of morphological and syntactic features specific to Indian languages.

\begin{table*}[hbt]
\renewcommand{\thetable}{4}
\label{tab:tokenizer_comparison}
\scriptsize
\centering
\begin{tabular}{c c c c c c c c}
\toprule
\textbf{Tokenizer Name} & \textbf{Eval Loss} & \textbf{Context} & \textbf{Completeness} & \textbf{Creativity} & \textbf{Fluency} & \textbf{Grammar} & \textbf{Overall} \\
\midrule
\multicolumn{8}{l}{\textbf{Hindi}} \\
\midrule
Sarvam   & \ColorEvalLossCell{0.518}{red} & \ColorGeneralCell{7.734}{red} & \ColorGeneralCell{7.783}{red} & \ColorGeneralCell{7.806}{red} & \ColorGeneralCell{8.554}{red} & \ColorGeneralCell{8.912}{red} & \ColorGeneralCell{8.158}{red} \\
SUTRA    & \ColorEvalLossCell{0.522}{red} & \ColorGeneralCell{7.548}{red} & \ColorGeneralCell{7.449}{red} & \ColorGeneralCell{7.584}{red} & \ColorGeneralCell{8.292}{red} & \ColorGeneralCell{8.875}{red} & \ColorGeneralCell{7.950}{red} \\
Tiktoken & \ColorEvalLossCell{0.149}{red} & \ColorGeneralCell{6.974}{red} & \ColorGeneralCell{7.106}{red} & \ColorGeneralCell{7.360}{red} & \ColorGeneralCell{7.889}{red} & \ColorGeneralCell{8.681}{red} & \ColorGeneralCell{7.602}{red} \\
\midrule
\multicolumn{8}{l}{\textbf{Marathi}} \\
\midrule
Sarvam   & \ColorEvalLossCell{0.645}{teal} & \ColorGeneralCell{7.245}{teal} & \ColorGeneralCell{7.407}{teal} & \ColorGeneralCell{7.553}{teal} & \ColorGeneralCell{8.106}{teal} & \ColorGeneralCell{8.723}{teal} & \ColorGeneralCell{7.807}{teal} \\
SUTRA    & \ColorEvalLossCell{0.627}{teal} & \ColorGeneralCell{7.523}{teal} & \ColorGeneralCell{7.162}{teal} & \ColorGeneralCell{7.483}{teal} & \ColorGeneralCell{8.012}{teal} & \ColorGeneralCell{8.724}{teal} & \ColorGeneralCell{7.781}{teal} \\
Tiktoken & \ColorEvalLossCell{0.167}{teal} & \ColorGeneralCell{7.014}{teal} & \ColorGeneralCell{6.742}{teal} & \ColorGeneralCell{7.137}{teal} & \ColorGeneralCell{7.524}{teal} & \ColorGeneralCell{8.451}{teal} & \ColorGeneralCell{7.374}{teal} \\
\midrule
\multicolumn{8}{l}{\textbf{Bengali}} \\
\midrule
Sarvam   & \ColorEvalLossCell{0.569}{beige} & \ColorGeneralCell{7.507}{beige} & \ColorGeneralCell{7.645}{beige} & \ColorGeneralCell{7.693}{beige} & \ColorGeneralCell{8.420}{beige} & \ColorGeneralCell{8.816}{beige} & \ColorGeneralCell{8.016}{beige} \\
SUTRA    & \ColorEvalLossCell{0.608}{beige} & \ColorGeneralCell{7.614}{beige} & \ColorGeneralCell{7.374}{beige} & \ColorGeneralCell{7.595}{beige} & \ColorGeneralCell{8.212}{beige} & \ColorGeneralCell{8.845}{beige} & \ColorGeneralCell{7.928}{beige} \\
Tiktoken & \ColorEvalLossCell{0.135}{beige} & \ColorGeneralCell{7.118}{beige} & \ColorGeneralCell{6.989}{beige} & \ColorGeneralCell{7.358}{beige} & \ColorGeneralCell{7.778}{beige} & \ColorGeneralCell{8.614}{beige} & \ColorGeneralCell{7.572}{beige} \\
\bottomrule
\end{tabular}
\caption{\small{Comparison of tokenizers across Hindi, Marathi, and Bengali for model with 6 layers, 8 attention heads, 512 hidden embeddings.}}
\label{tab:tokenizer-comparison}
\end{table*}
\subsection{Models trained on translated data have lower inference evaluations}

The 54M model with Sarvam tokenizer was chosen for inference evaluations on translated datasets and compared with those trained on synthetic data. We observe that these models have higher evaluation loss for same training duration and have poorer inference evaluation scores , suggesting that translated data quality is lower compared to synthetic. For e.g., the Hindi model has an overall score of 6.39 against 8.16 observed for model trained with synthetic data (Appendix F). Reduced performance, upon using translated data, was also noticed in one previous report \cite{boughorbel2024improvinglanguagemodelstrained}.
This could be explained using the following viewpoints: (1) Cultural biases: Source data culture transfers to target languages, including elements like foreign names that prevent models from generating culturally appropriate content \cite{holmstrom-etal-2023-bridging}. (2) Grammatical and style issues: Languages express similar concepts with different strucutures and conventions, and often translations fail to adapt these nuances, producing unnatural text in the target language \cite{zhang-toral-2019-effect}. (3) Lastly, higher evaluation losses suggest difficulty in next token prediction which could stem from the noise introduced through the translation process, that complicates the task compared to working with original text \cite{boughorbel2024improvinglanguagemodelstrained}.

\subsection{Tokenizer analysis with focus on language complexity }

We use a dual-perspective approach to quantify the linguistic complexity of Hindi, Bengali, and Marathi, revealing their intrinsic features for evaluating tokenization strategies.
\subsubsection{Information-Theoretic Analysis}
To evaluate tokenization quality and language complexity, we computed Rényi entropy \cite{zouhar-etal-2023-tokenization} information-theoretic measure of uncertainty and diversity in the tokenized distributions across the training corpora for each of the three languages using both Sarvam and SUTRA tokenizers. Rényi entropy provides a parameterized framework for quantifying information content in tokenized distributions, with parameter $\alpha$ controlling the sensitivity to rare versus common tokens.
Table 5 presents our findings:
\begin{table}[h]
    \centering
    \begin{tabular}{lccc}
        \toprule
        \textbf{Tokenizer}& \textbf{Hindi}& \textbf{Bengali}& \textbf{Marathi}\\
        \midrule
        Sarvam   & 6.2852 & 6.3579 & 6.5449 \\
        SUTRA    & 7.1530 & 7.4135 & 7.7620 \\
        \bottomrule
    \end{tabular}
    \caption{Rényi entropy (\(\alpha = 2.5\)) for Hindi, Bengali, and Marathi using Sarvam and SUTRA tokenizers.}
    \label{tab:renyi_entropy}
\end{table}

Our analysis reveals consistent patterns across tokenizers and languages. Marathi consistently exhibits the highest entropy values, which suggests that Marathi may possess a more complex morphological structure or greater variability in token-level patterns, necessitating a more diverse set of tokens for accurate representation. This could be the reason behind Marathi's overall lower evaluation scores.

The choice of tokenizer significantly influences entropy distributions. With Sarvam, we observe lower entropy values across all languages, indicating more concentrated probability mass in token distributions. This suggests Sarvam's design achieves more compact tokenization by capturing efficient subword structures for Indian languages. Conversely, SUTRA's higher entropy values point to a more diverse tokenization strategy, potentially offering richer representational capacity at the cost of increased vocabulary complexity. This could explain why models using Sarvam consistently outperform those with SUTRA.

To examine how entropy varies with the $\alpha$ parameter, we computed values at different levels ($\alpha$ = 0.5, 1.0, 2.0). At $\alpha$ = 0.5, which emphasizes rare tokens, Marathi showed the highest entropy (SUTRA: 10.69, Sarvam: 11.06). At $\alpha$ = 1.0 (Shannon entropy), the languages demonstrated moderate convergence, though Marathi maintained higher values. These consistent patterns across $\alpha$ values confirm robust differences in tokenization complexity among these languages.
\subsubsection{Morph Score}
To complement our information-theoretic approach, we evaluated morphological fidelity, on the words used for analyzing Rényi entropy, using MorphScore, which quantifies alignment between tokenizer outputs and linguistic morphemes. A 'morpheme' is the smallest unit of language with meaning, serving as a basic building block for words. Following the methodology established by \citeauthor{arnett2024languagemodelsperformworse} \citeyearpar{arnett2024languagemodelsperformworse}, we constructed morphologically-annotated evaluation sets for each language.
\begin{table}[h]
    \centering
    \begin{tabular}{lcc}
        \toprule
        \textbf{Language} & \textbf{SUTRA}& \textbf{Sarvam} \\
        \midrule
        Hindi    & 0.7268 & 0.7276 \\
        Bengali  & 0.3002 & 0.3194 \\
        Marathi  & 0.6671 & 0.6620 \\
        \bottomrule
    \end{tabular}
    \caption{MorphScore evaluation results comparing SUTRA and Sarvam tokenizers across three Indic languages. Higher scores indicate better alignment with morphological boundaries.}
    \label{tab:morphscores}
\end{table}
The MorphScore results in Table 6 reveal several interesting patterns. First, Sarvam achieves marginally higher MorphScore values for Hindi and Bengali, over SUTRA, suggesting better preservation of morphological boundaries for these languages in its tokenization strategy. Second, we observe substantial variation in absolute MorphScore values across languages, with Bengali showing markedly lower scores (by 50 \%) compared to Hindi and Marathi. This stark difference suggests current tokenization approaches may not optimally capture Bengali's morphological structures. According to \citeauthor{arnett2024languagemodelsperformworse} \citeyearpar{arnett2024languagemodelsperformworse}'s analysis a higher MorphScore indicates a better aligned tokenizer and consequently better model performance, which in our case is broken by the Bengali models outperforming Marathi. Hence, other mechanistic factors like Rényi entropy may have greater language specific impact, necessitating more research in this area.

\subsubsection{Cross-Linguistic Complexity Analysis}

The above analyses reveal interesting insights into language complexity. While Marathi shows higher complexity in terms of token distribution and processing requirements (highest Rényi entropy values), Bengali presents unique challenges in morphological boundary recognition (lowest MorphScore). Hindi consistently shows moderate values across both metrics, potentially explaining its relatively efficient tokenization and consequently the best performance in our model evaluations. Also, Sarvam shows consistently lower entropy and higher morphological alignment (except for Marathi) and thus better performance which is again corroborated by the inference scores.

\citeauthor{bafna-zabokrtsky-2022-subword} \citeyearpar{bafna-zabokrtsky-2022-subword} have previously discussed Marathi being more agglutinative than Hindi, thus allowing suffix stacking with boundary changes. They show that a Marathi token might combine verb, nominalizing morpheme and case marker, while Hindi often separates these into individual tokens. This aligns with our empirical observations of model performance, where Marathi models typically required 20-30 $\% $ more parameters to achieve comparable performance levels to Hindi and Bengali models. The consistency between theoretical complexity measures and empirical model behavior provides measurable validation for our hypothesis of using SLMs to evaluate language complexity.

It is crucial to note that these complexity measures are conditional on the chosen tokenization strategy. As demonstrated previosuly \cite{arnett2024languagemodelsperformworse}, no language is inherently harder or easier for a language model to learn based solely on its morphological topology or tokenization strategy. The variations we observe in model performance likely stem from the interaction between intrinsic linguistic properties and practical factors such as dataset size and quality.

\section{Conclusion}

We demonstrated the TinyStories paradigm's effectiveness for Indian languages, showing SLMs with just 5-50 million parameters can generate coherent outputs. Our largest model ($\approx 150M$ parameters) achieves 90 \% of GPT-4o's performance ($\approx 8$ vs $\approx 9$ on overall evaluations), despite being $10^6$ times smaller.

Further Regional TinyStories provides a novel approach for comparing language complexities and tokenizer efficiencies based on our inference evaluation framework. We find that Hindi models using Sarvam perform the best. Its observed that Bengali favors creativity, Hindi has slightly weaker context-completeness links, and Marathi uniquely associates context with grammar, suggesting distinct learning capabilities. Through comprehensive evaluation, we established that Sarvam and SUTRA tokenizers outperform alternatives like Tiktoken —the first study comparing regional tokenizers through inference evaluations rather than token fertility metrics. Our mechanistic analysis points toward higher entropy in tokenization behind the lower overall performance for Marathi compared to Hindi and Bengali, suggesting potential differences in how tokenizers deal with language complexity.

Our work demonstrates that effective Indian language modeling may not require massive architectures but rather quality, focused datasets, potentially democratizing language technology for underrepresented languages.

\clearpage

\newpage
\section{Limitation}

\begin{enumerate}
    \item Due to computational resource constraint we could not run the same model several times for gathering more statistics, even though for each model run, 3000 stories were generated for inference evaluations.imilarly we could not translate English to Marathi.
    \item We do not incorporate human-in-the-loop evaluations of the SLM generated stories. The use of LLM as judge is still an exploratory topic. Potential biases and reliability of evaluation is an active field of research \cite{chen-etal-2024-humans}.
    \item While our 54M-parameter models achieve decent scores across Hindi, Marathi, and Bengali, we observed consistent patterns where context scores (7.2-7.7) lag behind fluency scores (8.1-8.6). This may be explained by previous research \cite{peng2022guidingneuralstorygeneration,peng2023inferringreaderguidingautomated} about neural models' challenges with entity-relationship tracking and deeper causal consistency. 
    \item WeightWatcher analysis (Appendix G) revealed under-training in our models, suggesting potential benefits from additional training epochs and targeted regularization.
    \item We did not visualize and analyze the attention and activation maps
of the models, and show how they relate to the generation process and the story content for different languages. This may shed more light on the observed variety in learning different linguistic attributes in case of Hindi or Marathi models.
    \item Future work should explore hybrid architectures combining our models' strong fluency with explicit entity-relationship tracking mechanisms, potentially bridging the gap between statistical pattern recognition and human-like narrative understanding in Indian language story generation.

\end{enumerate}

\section*{Impact Statement}

Our work in generating children's stories in Indian regional languages presents both significant opportunities and challenges for educational accessibility and cultural preservation. We intend to release this as open weight models for deployment across diverse environments from edge devices to cloud infrastructure. While this technology could democratize access to children's literature in underserved languages and support early childhood literacy in resource-constrained environments, it raises important considerations about cultural authenticity and content quality. The system's ability to generate low-cost, scalable content could help address the scarcity of children's literature in many languages, particularly beneficial for rural areas with limited publishing infrastructure. However, this advancement necessitates careful consideration of cultural nuances, content moderation, and the preservation of regional storytelling traditions. To ensure responsible deployment, we recommend implementing robust review mechanisms involving language experts, establishing clear guidelines for cultural appropriateness, and developing metrics to measure educational impact. The technology should complement rather than replace traditional storytelling, working in concert with local educators and cultural experts to maintain authenticity while leveraging the benefits of AI-assisted content generation.

\textbf{Acknowledgement}

We gratefully acknowledge the generous support from TensorDock (\url{https://tensordock.com}), whose \$1,500 research grant enabled the computational experiments in this work. Their cloud computing infrastructure was instrumental in training and evaluating our regional language models. 

\textbf{Code and model repository}

We publicly release all our codes and datasets used for the experiments described in this work as well as selected trained models, to allow for external validation and reproducibility :\\

\begin{enumerate}
    \item Code : \url{https://github.com/VizuaraAI/Tiny-Stories-Regional}
    \item Code : \url{https://github.com/nirvan840/Vizuara-TinyStories-Regional}
    \item Dataset and models : \url{https://huggingface.co/TinyStories-Regional}
\end{enumerate}

\newpage
\nocite{langley00}

\bibliography{example_paper}
\bibliographystyle{icml2025}

\newpage
\appendix
\onecolumn
{\huge \textbf{Appendix}}

\section{Renyi Entropy of Hindi, Marathi, and Bengali with SUTRA and Sarvam Tokenizers}

\begin{table}[h]
\centering
\begin{tabular}{lcccc}
\toprule
Language & Tokenizer & $\alpha=0.5$ & $\alpha=1.0$ & $\alpha=2.0$ \\
\midrule
Hindi & SUTRA & 9.77 & 8.67 & 7.49 \\
Hindi & Sarvam & 10.07 & 8.51 & 6.75 \\
\midrule 
Marathi & SUTRA & 10.69 & 9.37 & 8.15 \\
Marathi & Sarvam & 11.06 & 9.40 & 7.21 \\
\midrule 
Bengali & SUTRA & 9.82 & 8.79 & 7.72 \\
Bengali & Sarvam & 9.91 & 8.63 & 6.91 \\
\bottomrule
\end{tabular}
\caption{Rényi entropy of Hindi, Marathi, and Bengali with SUTRA and Sarvam tokenizers}
\label{tab:renyi_entropy}
\end{table}


When examining these metrics at different values of $\alpha$ (0.5, 1.0, 2.0), we observe consistent patterns that illuminate different aspects of language complexity:

\begin{enumerate}
    \item At $\alpha=0.5$, emphasizing rare tokens, Marathi shows the highest entropy (SUTRA: 10.69, Sarvam: 11.06), suggesting greater diversity in its rare token distributions.
    \item At $\alpha=1.0$ (Shannon entropy), all languages show moderate convergence, though Marathi maintains higher values (SUTRA: 9.37, Sarvam: 9.40).
    \item At $\alpha=2.0$, emphasizing common tokens, the differences between languages become less pronounced, though the relative ordering remains consistent.
\end{enumerate}

According to \citeauthor{arnett2024languagemodelsperformworse} \citeyearpar{arnett2024languagemodelsperformworse}, agglutinative languages have higher Rényi entropy compared to fusional languages. A study comparing Hindi, Marathi, and Bengali notes that Marathi's agglutinative structure creates more complex inflectional patterns, requiring distinct stemming strategies for information retrieval tasks. Bengali's simpler fusional morphology contrasts with Marathi's suffix-heavy word formation \cite{10.1145/1838745.1838748}.

\section{Statistical Analysis of Inference Results}

More details are provided for the inference evaluations for the 54M model that used Sarvam tokenizer.

\subsubsection{Distribution Analysis of Evaluation Metrics}

We conducted a comprehensive statistical analysis of the 3,000 stories generated by our Small Language Models (SLMs) for each of the three target languages: Hindi, Bengali, and Marathi, using Sarvam tokenizer. This analysis provides deeper insights into the performance characteristics of our models across various evaluation dimensions.

\subsubsection{Distributional Characteristics}

The evaluation scores for all three languages exhibit distinct distributional patterns that reveal important aspects of model behavior (Figures 4-9). Below, we summarize key statistical properties observed across languages and metrics.

\begin{table}[h]
\centering
\caption{Statistical Properties of Evaluation Metrics Across Languages for 54M models}
\label{tab:statistical_properties}
\begin{tabular}{llccc}
\hline
\textbf{Language} & \textbf{Metric} & \textbf{Mean} & \textbf{Median} & \textbf{Std Dev} \\
\hline
Hindi & Context Awareness & 7.73 & 8.00 & 1.01 \\
 & Completeness & 7.78 & 8.00 & 0.86 \\
 & Grammar & 8.91 & 9.00 & 0.34 \\
 & Fluency & 8.55 & 9.00 & 0.56 \\
 & Creativity & 7.81 & 8.00 & 0.58 \\
 & Overall & 7.79 & 8.00 & 0.52 \\
\hline
Marathi & Context Awareness & 7.25 & 8.00 & 1.18 \\
 & Completeness & 7.41 & 7.00 & 0.87 \\
 & Grammar & 8.72 & 9.00 & 0.50 \\
 & Fluency & 8.11 & 8.00 & 0.64 \\
 & Creativity & 7.55 & 8.00 & 0.69 \\
 & Overall & 7.50 & 8.00 & 0.67 \\
\hline
Bengali & Context Awareness & 7.51 & 8.00 & 1.11 \\
 & Completeness & 7.64 & 7.00 & 0.85 \\
 & Grammar & 8.82 & 9.00 & 0.42 \\
 & Fluency & 8.42 & 8.00 & 0.59 \\
 & Creativity & 7.69 & 8.00 & 0.59 \\
 & Overall & 7.68 & 8.00 & 0.57 \\
\hline
\end{tabular}
\end{table}

\subsubsection{Cross-Linguistic Performance Patterns}

Our analysis reveals several significant cross-linguistic patterns that provide insights into both model behavior and inherent language characteristics:

\begin{enumerate}
    \item \textbf{Hierarchical Emergence of Capabilities}: Across all three languages, we observe a consistent hierarchy in performance metrics, with grammar consistently achieving the highest scores (Hindi: 8.91, Marathi: 8.72, Bengali: 8.82), followed by fluency (Hindi: 8.55, Marathi: 8.11, Bengali: 8.42), completeness (Hindi: 7.78, Marathi: 7.41, Bengali: 7.64), and context awareness (Hindi: 7.73, Marathi: 7.25, Bengali: 7.51). This pattern aligns with the developmental progression observed in the original TinyStories research, suggesting that grammatical competence emerges earlier than contextual understanding, regardless of language.
    \item \textbf{Bimodal Distribution of Context Scores}: The violin plots reveal a distinctive bimodal distribution for context awareness scores across all three languages, with concentration of scores around 7 and 8-9 ranges. This bimodality suggests that stories tend to either achieve strong contextual coherence or struggle with maintaining context throughout the narrative, with relatively few stories falling in the intermediate range. This pattern is evident across all three languages but varies in intensity.
    \item \textbf{Consistency in Grammar Scores}: Grammar scores exhibit the lowest standard deviation across all languages (Hindi: 0.34, Marathi: 0.50, Bengali: 0.42), indicating that once basic grammatical competence is achieved, it remains relatively stable across generated stories. The narrow distribution of grammar scores visible in the violin plots demonstrates the models' tendency to consistently produce grammatically correct text.
    \item \textbf{Language-Specific Performance Differences}: Hindi outperforms both Bengali and Marathi across most metrics, with the most substantial advantage in grammar (Hindi: 8.91 vs. Marathi: 8.72) and fluency (Hindi: 8.55 vs. Marathi: 8.11). Marathi consistently shows lower performance across all metrics. This finding is particularly noteworthy given the relationship with Rényi entropy values discussed in the paper, suggesting that languages with higher entropy measures may present greater challenges for coherent text generation.
\end{enumerate}

\newpage
\subsection{Relationship Between Model Architecture and Evaluation Metrics}

To understand how different architectural choices affect specific linguistic capabilities, we conducted correlation analyses between model parameters and evaluation metrics.

\subsubsection{Parameter Efficiency Across Languages}

Tables 1-3 in the main text illustrate the relationship between model parameter count and evaluation metrics for each language. Several key observations emerge:

\begin{enumerate}
    \item \textbf{Divergent Scaling Patterns}: While all languages benefit from increased model size, the marginal improvements from scaling differ significantly. The comparable performance of similarly-sized models (54M parameters) across the three languages suggests that architectural scaling properties may be relatively consistent, though the absolute performance levels differ. Hindi demonstrates the strongest performance at this parameter range, followed by Bengali and then Marathi.
    \item \textbf{Optimal Parameter Allocation}: The inflection point in the performance-parameter curve occurs consistently around 54M parameters across all three languages, with minimal improvements beyond this threshold. However, the specific distribution of these parameters (between embedding dimension and layer depth) that yields optimal performance varies by language. Bengali achieves optimal performance with a balanced 512/6 configuration, while Hindi benefits more from increased width (1024/7) than depth. 
    \item \textbf{Parameter Elasticity by Metric}: Different evaluation metrics show varying sensitivity to parameter scaling. Grammar scores demonstrate the lowest elasticity (average 12\% improvement from 4.46M to 153M parameters across languages), while context awareness shows the highest (average 33\% improvement). This supports our hypothesis regarding the hierarchical emergence of capabilities, with grammatical competence requiring less model capacity than contextual understanding.
\end{enumerate}

\begin{figure}[H]
    \centering
    \includegraphics[width=0.6\textwidth]{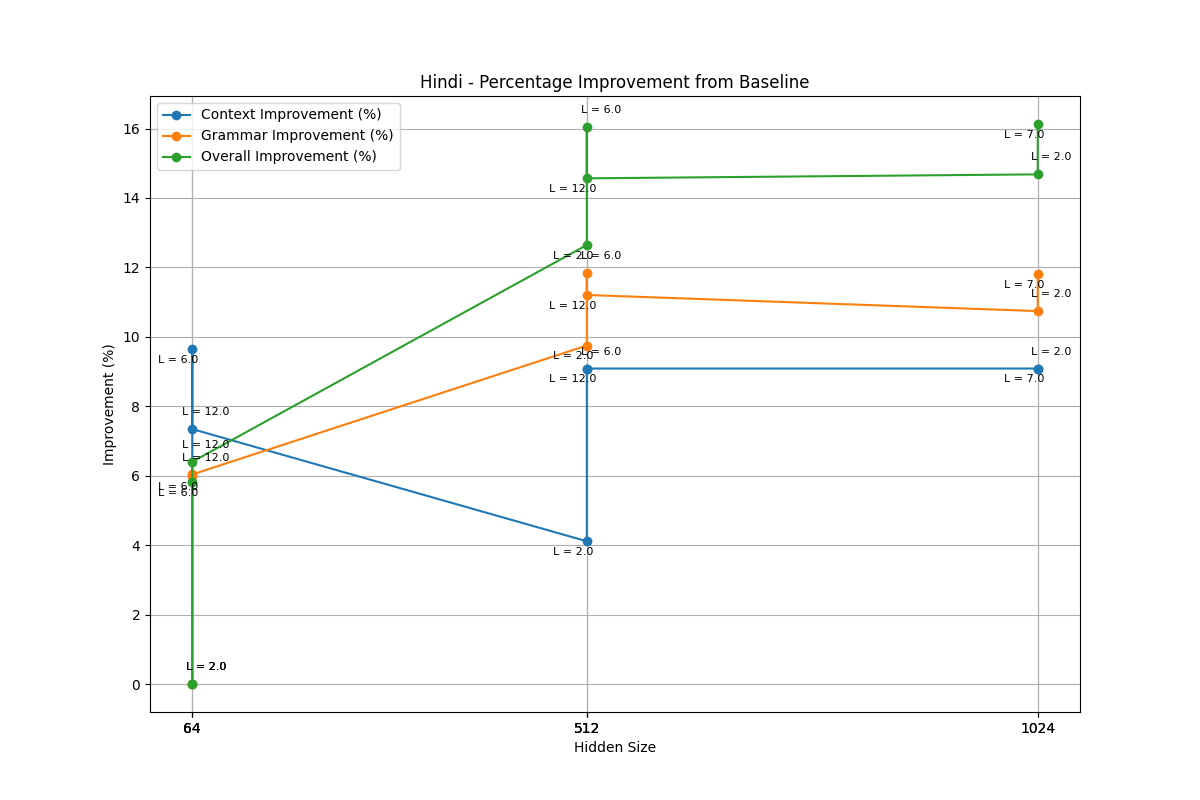}
    \caption{Percentage Improvement in Contextual Comprehension, Grammatical Accuracy, and Overall Performance from Baseline Scores across Hindi models detailed in Table 1 }
    \label{fig:hindi_improvement}
\end{figure}

\begin{figure}[ht]
    \centering
    \includegraphics[width=0.6\textwidth]{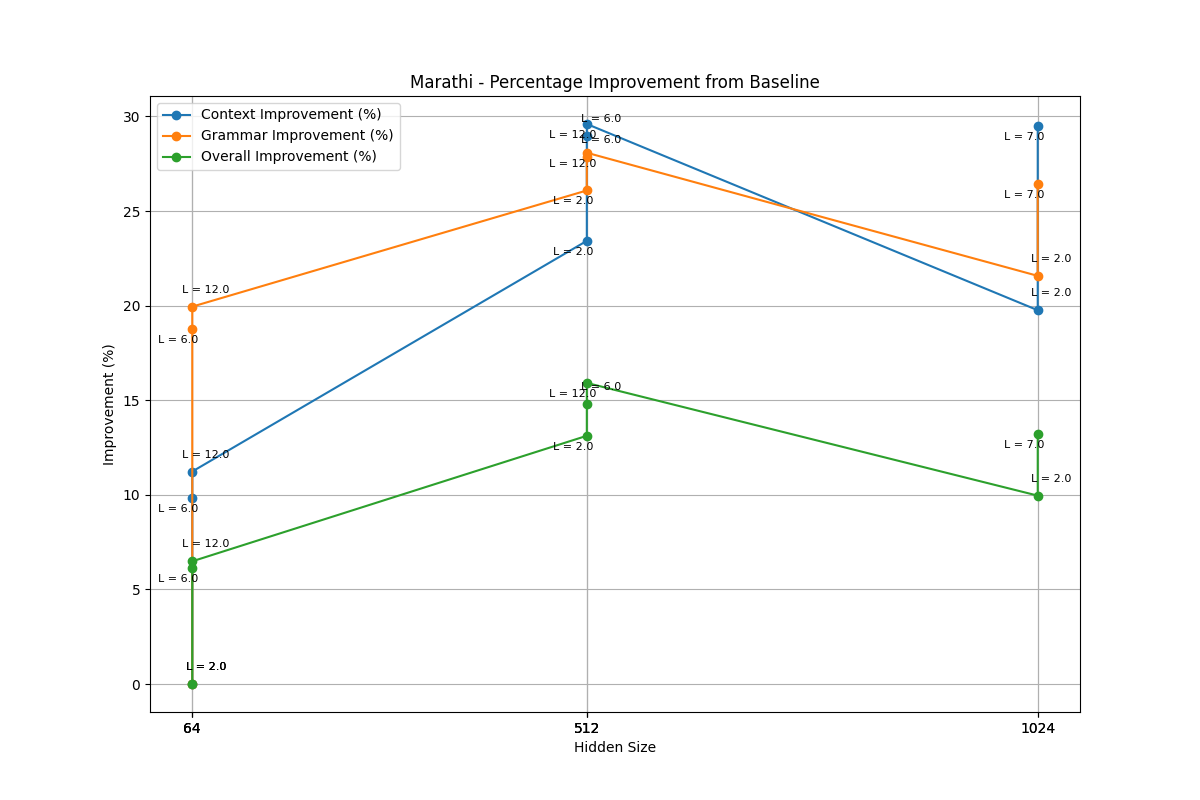}
    \caption{Percentage Improvement in Contextual Comprehension, Grammatical Accuracy, and Overall Performance from Baseline Scores across Marathi models detailed in Table 2 }
        \label{fig:marathi_improvement}
\end{figure}

\begin{figure}[ht]
    \centering
    \includegraphics[width=0.6\textwidth]{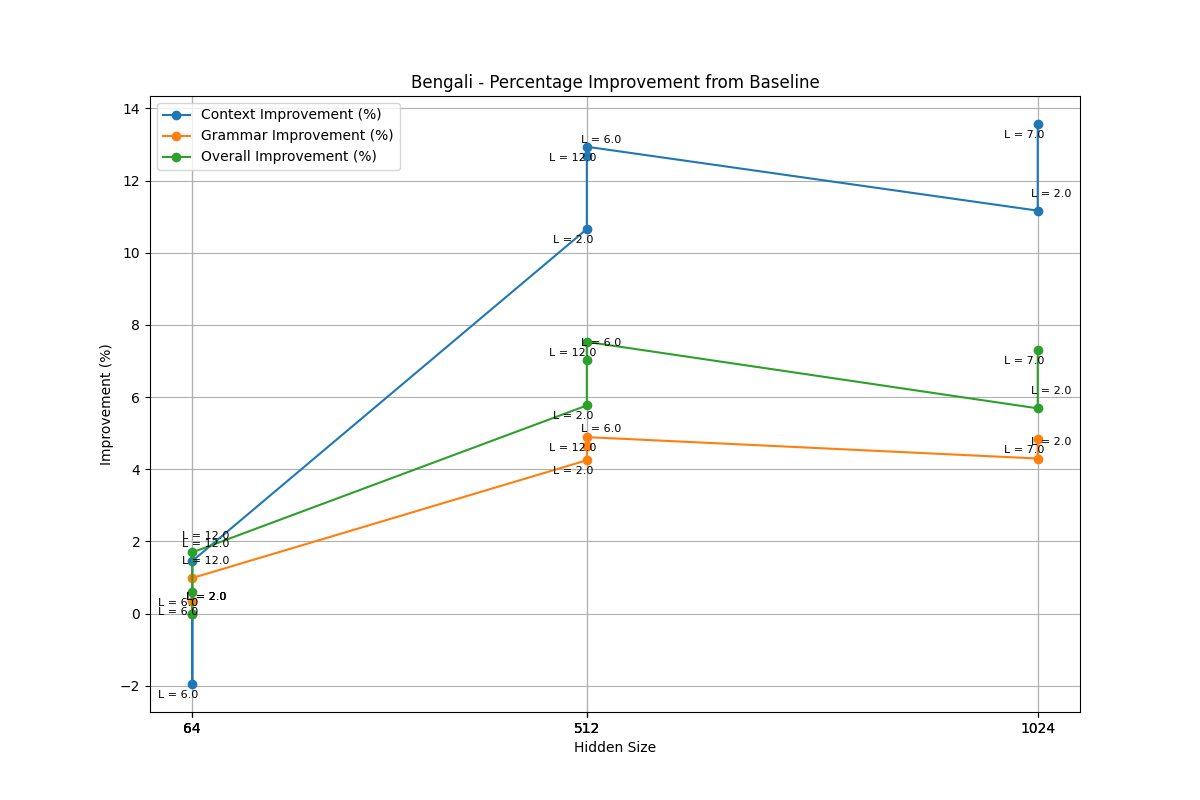}
    \caption{Percentage Improvement in Contextual Comprehension, Grammatical Accuracy, and Overall Performance from Baseline Scores across Bengali models detailed in Table 3 }
        \label{fig:bengali_improvement}
\end{figure}
\FloatBarrier

\subsection{Correlation Analysis Between Metrics}

\paragraph{Hindi}

The evaluation metrics in Hindi short stories demonstrated significant inter-correlations. The strongest association was observed between creativity and overall quality assessment (\textit{r} = 0.73, \textit{t}(2998) = 58.48, \textit{p} $<$ .001), indicating that creative elements substantially influenced holistic quality perceptions. Grammar and overall quality demonstrated a robust positive relationship (\textit{r} = 0.69, \textit{t}(2998) = 52.20, \textit{p} $<$ .001), suggesting grammatical accuracy significantly contributed to quality judgments. Notably, completeness and fluency exhibited a strong correlation (\textit{r} = 0.72, \textit{t}(2998) = 56.81, \textit{p} $<$ .001), indicating narrative completeness typically accompanied smooth readability. The weakest relationship was identified between context awareness and completeness (\textit{r} = 0.35, \textit{t}(2998) = 20.46, \textit{p} $<$ .001), suggesting these constructs captured distinct dimensions of narrative quality.

\paragraph{Bengali}

Analysis of Bengali short stories revealed similar correlation patterns, with creativity and overall quality showing the highest correlation coefficient (\textit{r} = 0.80, \textit{t}(2998) = 73.01, \textit{p} $<$ .001). This suggests that creative expression was the predominant factor in quality assessment. Grammar and overall quality maintained a strong positive relationship (\textit{r} = 0.71, \textit{t}(2998) = 55.20, \textit{p} $<$ .001), highlighting the importance of grammatical precision. Completeness and fluency demonstrated substantial correlation (\textit{r} = 0.77, \textit{t}(2998) = 66.08, \textit{p} $<$ .001), reinforcing the connection between narrative coherence and reading experience observed across languages. Context awareness and completeness displayed a moderate correlation (\textit{r} = 0.39, \textit{t}(2998) = 23.19, \textit{p} $<$ .001), indicating these metrics evaluated partially distinct aspects of narrative construction.

\paragraph{Marathi}

In contrast to Hindi and Bengali, Marathi short stories exhibited their strongest correlation between context awareness and grammar (\textit{r} = 0.78, \textit{t}(2998) = 68.25, \textit{p} $<$ .001), suggesting a language-specific relationship between contextual appropriateness and grammatical structure. Creativity and overall quality maintained equivalent correlation strength (\textit{r} = 0.78, \textit{t}(2998) = 68.25, \textit{p} $<$ .001), consistent with patterns observed in the other languages. Completeness and fluency correlation remained robust (\textit{r} = 0.77, \textit{t}(2998) = 66.08, \textit{p} $<$ .001), indicating a consistent relationship across all three languages. The weakest association was observed between completeness and creativity (\textit{r} = 0.49, \textit{t}(2998) = 30.78, \textit{p} $<$ .001), suggesting these dimensions function more independently in Marathi narratives compared to Hindi and Bengali.

These findings reveal both cross-linguistic patterns and language-specific relationships between evaluation metrics, with implications for understanding quality assessment in Indic-language short stories. Given the large sample size (\textit{n} = 3000 per language), all correlations were statistically significant at \textit{p} $<$ .001, with the critical value for significance at this level being \textit{r} = 0.060.


\subsection{Comparative Analysis of Score Distributions}

\subsubsection{Distribution Variation Analysis}

Examining the standard deviations across metrics and languages provides insight into the consistency of model performance:

\begin{table}[h]
\centering
\caption{Standard Deviation Comparison Across Metrics and Languages}
\label{tab:std_dev_comparison}
\begin{tabular}{lcccc}
\hline
\textbf{Metric} & \textbf{Hindi} & \textbf{Marathi} & \textbf{Bengali} & \textbf{Average} \\
\hline
Context Awareness & 1.01 & 1.18 & 1.11 & 1.10 \\
Completeness & 0.86 & 0.87 & 0.85 & 0.86 \\
Grammar & 0.34 & 0.50 & 0.42 & 0.42 \\
Fluency & 0.56 & 0.64 & 0.59 & 0.60 \\
Creativity & 0.58 & 0.69 & 0.59 & 0.62 \\
Overall & 0.52 & 0.67 & 0.57 & 0.59 \\
\hline
Average & 0.65 & 0.76 & 0.69 & 0.70 \\
\hline
\end{tabular}
\end{table}

Notable patterns include:

\begin{enumerate}
    \item \textbf{Consistent Hierarchy of Variability}: Across all languages, Context Awareness shows the highest standard deviation (average: 1.10), indicating greater variability in the model's ability to maintain contextual coherence. Grammar consistently shows the lowest standard deviation (average: 0.42), suggesting that grammatical competence is more uniformly achieved once the model reaches sufficient capacity.
    \item \textbf{Language-Specific Consistency Patterns}: Marathi shows higher standard deviations across all metrics (average: 0.76) compared to Hindi (0.65) and Bengali (0.69), suggesting greater variability in performance. This is particularly evident in context awareness (Marathi: 1.18 vs. Hindi: 1.01) and overall scores (Marathi: 0.67 vs. Hindi: 0.52).
    \item \textbf{Form vs. Content Metrics}: Metrics related to linguistic form (grammar, fluency) consistently show lower standard deviations (0.42, 0.60) than those related to content (context, completeness, creativity) (1.10, 0.86, 0.62). This pattern suggests that form-related capabilities may develop more uniformly compared to content-related capabilities.
\end{enumerate}

\subsubsection{Consistency-Performance Relationship}

Examining the relationship between metric means and standard deviations reveals important patterns:

\begin{enumerate}
    \item \textbf{Inverse Relationship}: Across all languages, we observe an inverse relationship between mean scores and standard deviations. Metrics with higher means (like grammar) tend to have lower standard deviations, while metrics with lower means (like context awareness) show higher standard deviations. This pattern suggests that as performance on a particular aspect improves, consistency also increases.
    \item \textbf{Language-Specific Consistency}: All three languages show a moderate to strong negative correlation between means and standard deviations, with Marathi having the strongest inverse relationship (r = - 0.77), while both Bengali and Hindi show identical correlations (r = - 0.70). This negative correlation indicates that metrics with higher mean scores tend to have lower variability across all three languages, suggesting more consistent performance in areas where the models score higher.
    \item \textbf{Metric-Specific Patterns}: Grammar shows both the highest means and lowest standard deviations across all languages, suggesting that grammatical competence represents a "foundational" capability that is both strong and consistent once achieved. Context awareness, by contrast, shows lower means and higher standard deviations, indicating it may represent a more advanced capability that remains challenging even as models improve.
\end{enumerate}

\subsection{Performance Gap Analysis}

To better understand the relative strengths and weaknesses of the models across different languages, we analyze the gaps between different evaluation metrics:

\begin{table}[h]
\centering
\caption{Performance Gaps Between Metrics (Difference in Mean Scores)}
\label{tab:performance_gaps}
\begin{tabular}{lcccc}
\hline
\textbf{Metric Pair} & \textbf{Hindi} & \textbf{Marathi} & \textbf{Bengali} & \textbf{Average} \\
\hline
Grammar - Context Awareness & 1.18 & 1.47 & 1.31 & 1.32 \\
Grammar - Completeness & 1.13 & 1.31 & 1.18 & 1.21 \\
Grammar - Creativity & 1.10 & 1.17 & 1.13 & 1.13 \\
Grammar - Fluency & 0.36 & 0.61 & 0.40 & 0.46 \\
Fluency - Context Awareness & 0.82 & 0.86 & 0.91 & 0.86 \\
Fluency - Completeness & 0.77 & 0.70 & 0.78 & 0.75 \\
Fluency - Creativity & 0.74 & 0.56 & 0.73 & 0.68 \\
Context - Completeness & -0.05 & -0.16 & -0.13 & -0.11 \\
\hline
\end{tabular}
\end{table}

Key patterns include:

\begin{enumerate}
    \item \textbf{Consistent Gap Hierarchy}: Across all languages, the largest performance gap is between grammar and context awareness (average: 1.32), while the smallest gap among the major metric pairs is between context awareness and completeness (average: -0.11, with context scores actually lower than completeness in all languages). This consistent pattern suggests a universal hierarchy in how different linguistic capabilities develop in these models.
    \item \textbf{Language-Specific Gap Patterns}: Marathi shows notably larger gaps between grammar and other metrics (Grammar-Context: 1.47, Grammar-Completeness: 1.31) compared to Hindi and Bengali. This suggests that while Marathi models achieve reasonable grammar scores, they struggle more with contextual coherence and narrative completeness compared to models in other languages.
    \item \textbf{Form-Content Divide}: The substantial gaps between form-related metrics (grammar, fluency) and content-related metrics (context, completeness, creativity) highlight the models' stronger capabilities in producing structurally correct text versus semantically coherent narratives. This divide is most pronounced in Marathi and least evident in Hindi.
    \item \textbf{Grammar-Fluency Relationship}: The gap between grammar and fluency scores is significantly smaller (average: 0.46) than between grammar and other metrics, suggesting these capabilities may develop in tandem. This pattern holds across all three languages, though Marathi shows a larger grammar-fluency gap (0.61) compared to Hindi (0.36) and Bengali (0.40).
\end{enumerate}

\subsection{Statistical Significance Analysis}

To assess the significance of observed cross-linguistic differences, we analyzed the overall performance scores across the three languages:

\begin{table}[h]
\centering
\caption{Overall Performance Statistics by Language}
\label{tab:overall_stats}
\begin{tabular}{lcc}
\hline
\textbf{Language} & \textbf{Mean} & \textbf{Standard Deviation} \\
\hline
Hindi & 7.79 & 0.52 \\
Marathi & 7.50 & 0.67 \\
Bengali & 7.68 & 0.57 \\
\hline
\end{tabular}
\end{table}

These results suggest that:

\begin{enumerate}
    \item The performance differences between languages appear meaningful, with Hindi outperforming both Bengali and Marathi, and Marathi showing the lowest overall performance.
    \item The standard deviations indicate different levels of consistency across languages, with Hindi showing the most consistent performance (SD = 0.52) and Marathi showing the greatest variability (SD = 0.67).
    \item These performance differences align with the entropy analysis presented in the main paper, where Marathi exhibited higher Rényi entropy values (7.76) compared to Hindi (7.15) and Bengali (7.41), suggesting a potential relationship between tokenization complexity and generation performance.
\end{enumerate}

\subsection{Final comments}

Our statistical analysis reveals complex relationships between tokenization strategies, linguistic properties, and generation performance across Hindi, Marathi, and Bengali. The consistent hierarchy of capabilities (grammar $>$ fluency $>$ completeness $>$ context) across all three languages suggests universal aspects of language model development, while significant cross-linguistic differences in absolute performance point to the importance of language-specific optimization.

The performance metrics for comparable model architectures (53-54M parameters) across the three languages show Hindi achieving the strongest overall results (7.79), followed by Bengali (7.68) and Marathi (7.50). This aligns with the Rényi entropy analysis presented in the main paper, suggesting that languages with higher entropy values may present greater challenges for coherent text generation.

These findings underscore the value of the Regional TinyStories framework as both a practical approach to developing efficient language models for Indian languages and as an analytical tool for understanding comparative linguistic complexity. Future work should focus on exploring the relationship between tokenization strategies, morphological characteristics, and model performance to develop more comprehensive metrics for predicting language modeling difficulty across typologically diverse languages.

Figures 4-9 show the histograms and violin plots for the Hindi, Marathi and Bengali languages for 3000 samples for the following metrics: context awareness, completeness, grammar, fluency, creativity and overall score.  For each of the languages, the mean and the median scores for the different evaluation categories are provided by the red and the green dashed lines respectively on the subplots.

\begin{figure}[ht]
    \centering
    \begin{minipage}[b]{0.48\textwidth}
        \centering
        \includegraphics[width=\textwidth]{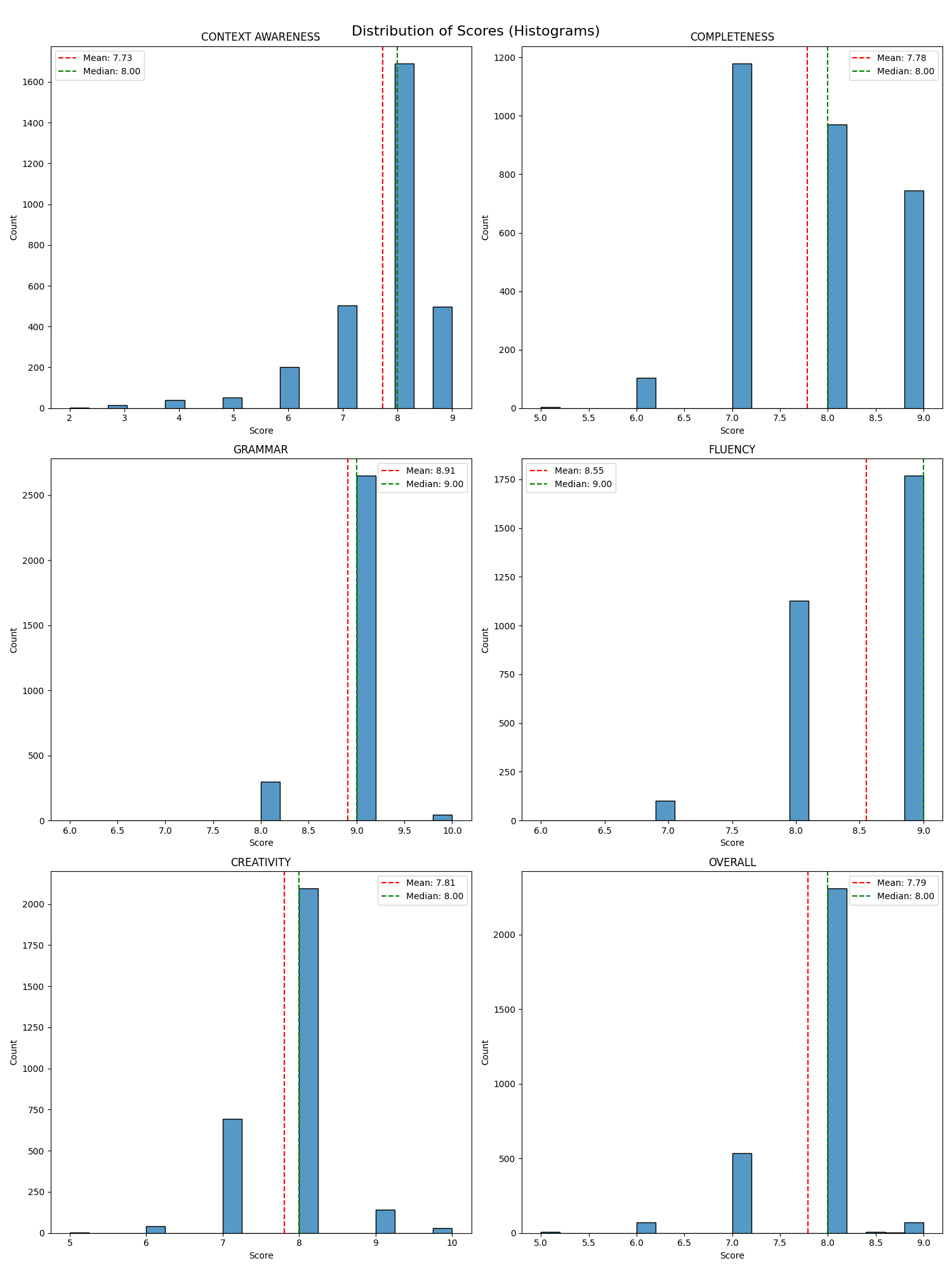}
        \caption{Hindi 54M inference score distribution, n = 3000}
        \label{fig:figure3}
    \end{minipage}
    \hfill
    \begin{minipage}[b]{0.48\textwidth}
        \centering
        \includegraphics[width=\textwidth]{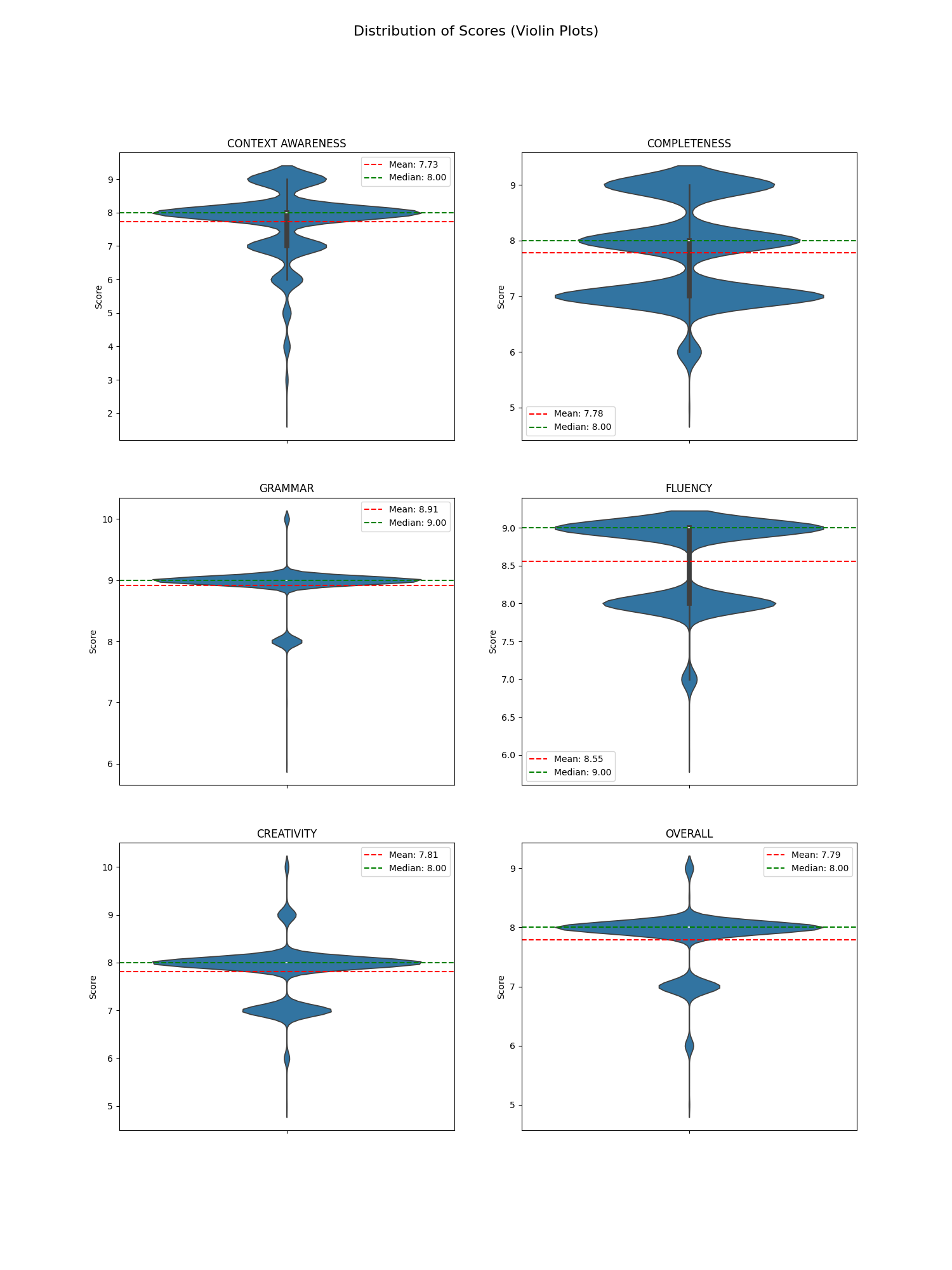}
        \caption{Hindi 54M inference score violin plots, n = 3000}
        \label{fig:figure4}
    \end{minipage}
\end{figure}
\begin{figure}[htbp]
\centering
    \begin{minipage}[t]{0.45\textwidth}
        \centering
        \includegraphics[width=0.9\textwidth]{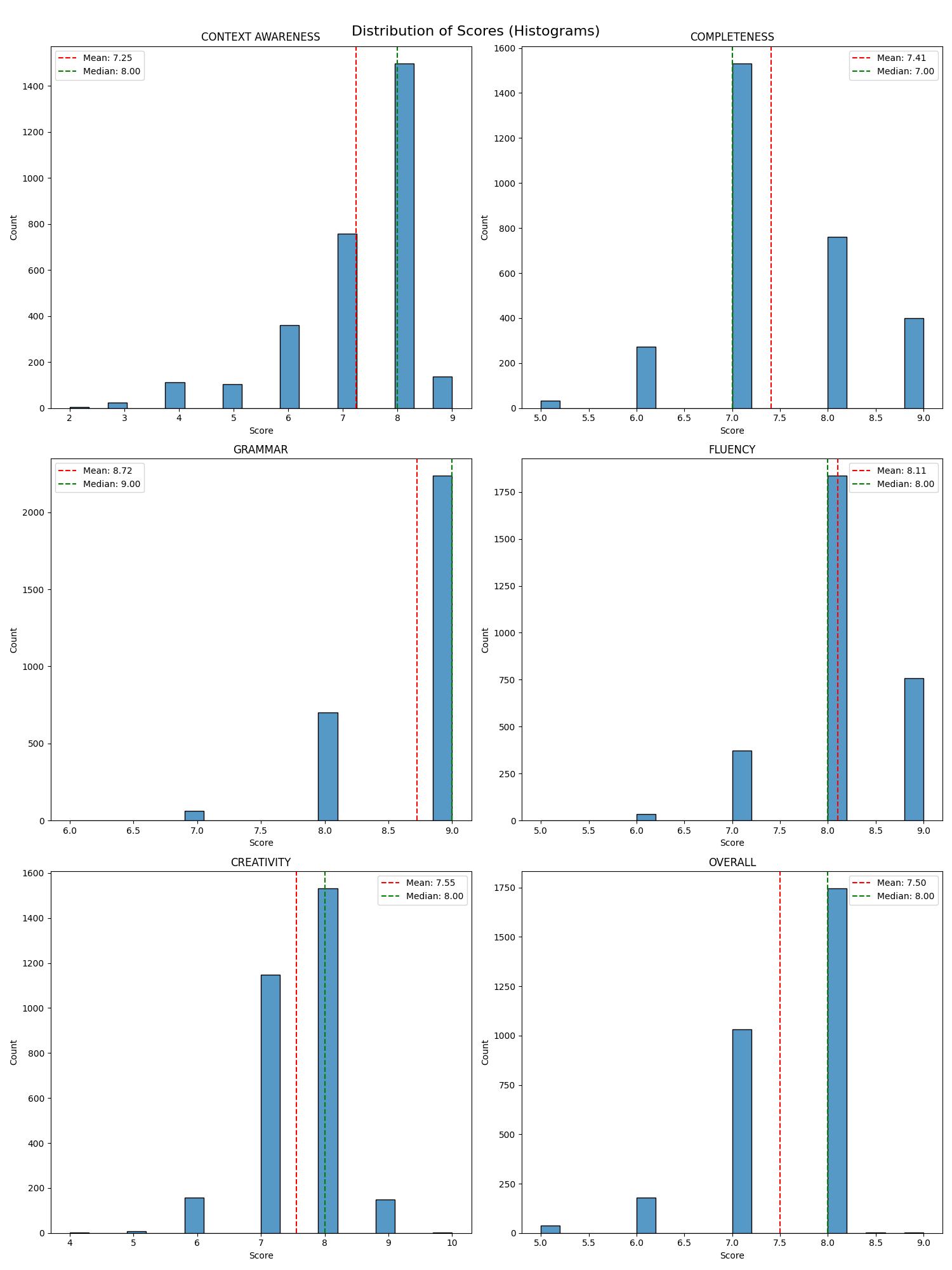}
        \caption{Marathi 54M inference score distribution, n=3000}
        \label{fig:marathi}
    \end{minipage}
    \hfill
    \begin{minipage}[t]{0.45\textwidth}
        \centering
        \includegraphics[width=0.9\textwidth]{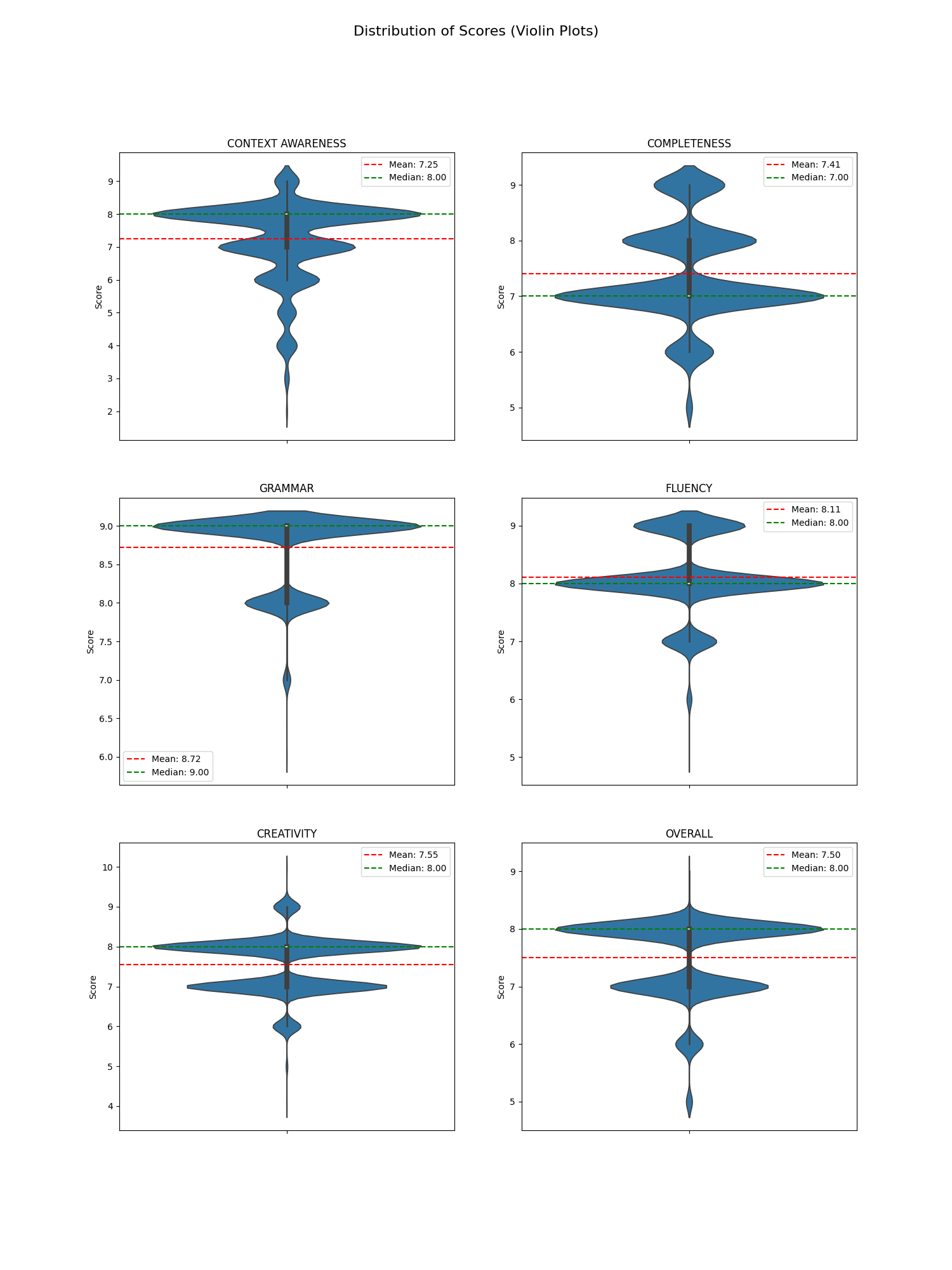}
        \caption{Marathi 54M inference score violin plots, n=3000}
        \label{fig:bengali}
    \end{minipage}
\end{figure}
\begin{figure}[ht]
    \centering
    \begin{minipage}[b]{0.48\textwidth}
        \centering
        \includegraphics[width=\textwidth]{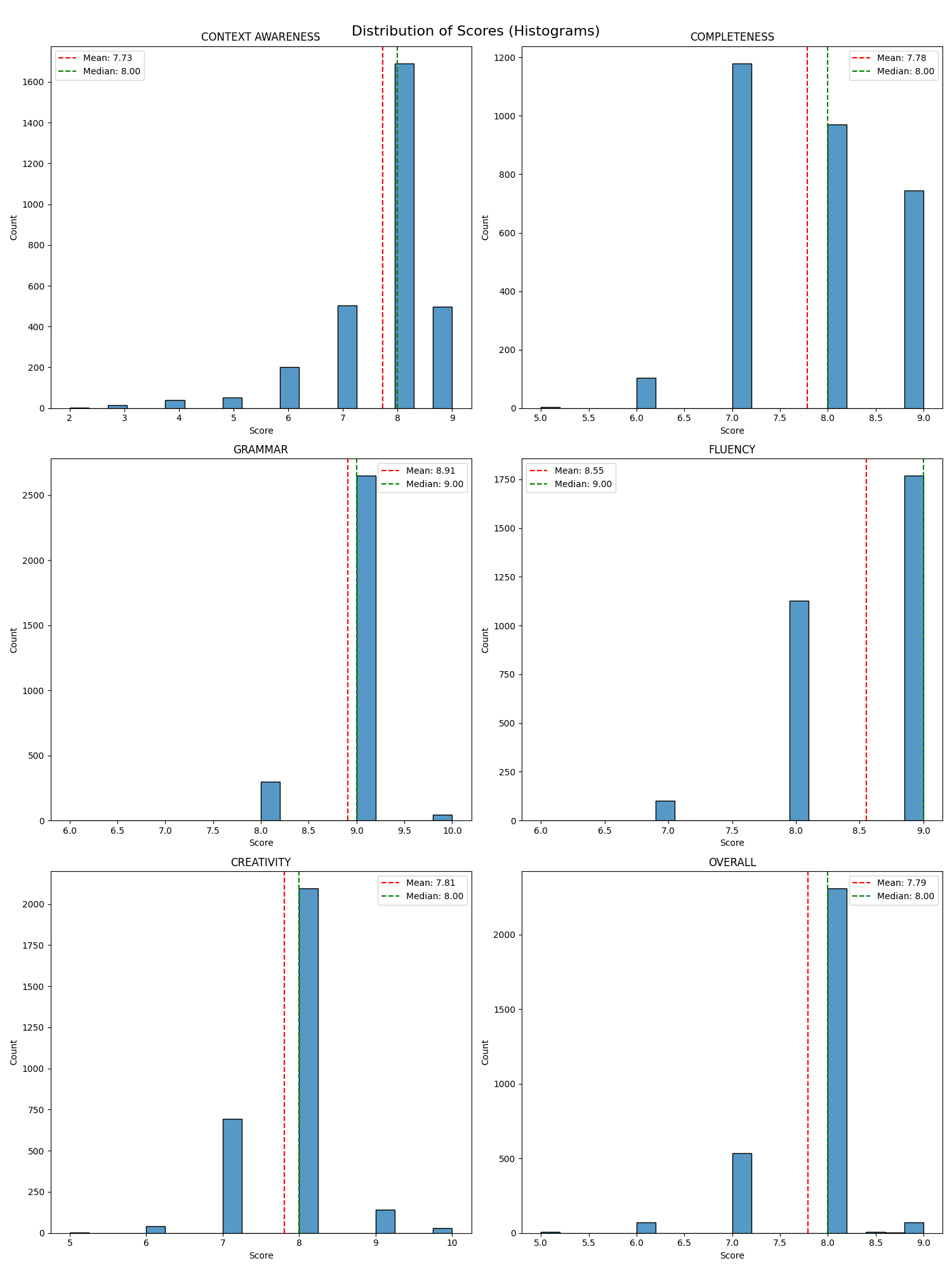}
        \caption{Bengali 54M inference score distribution, n = 3000}
        \label{fig:figure3}
    \end{minipage}
    \hfill
    \begin{minipage}[b]{0.48\textwidth}
        \centering
        \includegraphics[width=\textwidth]{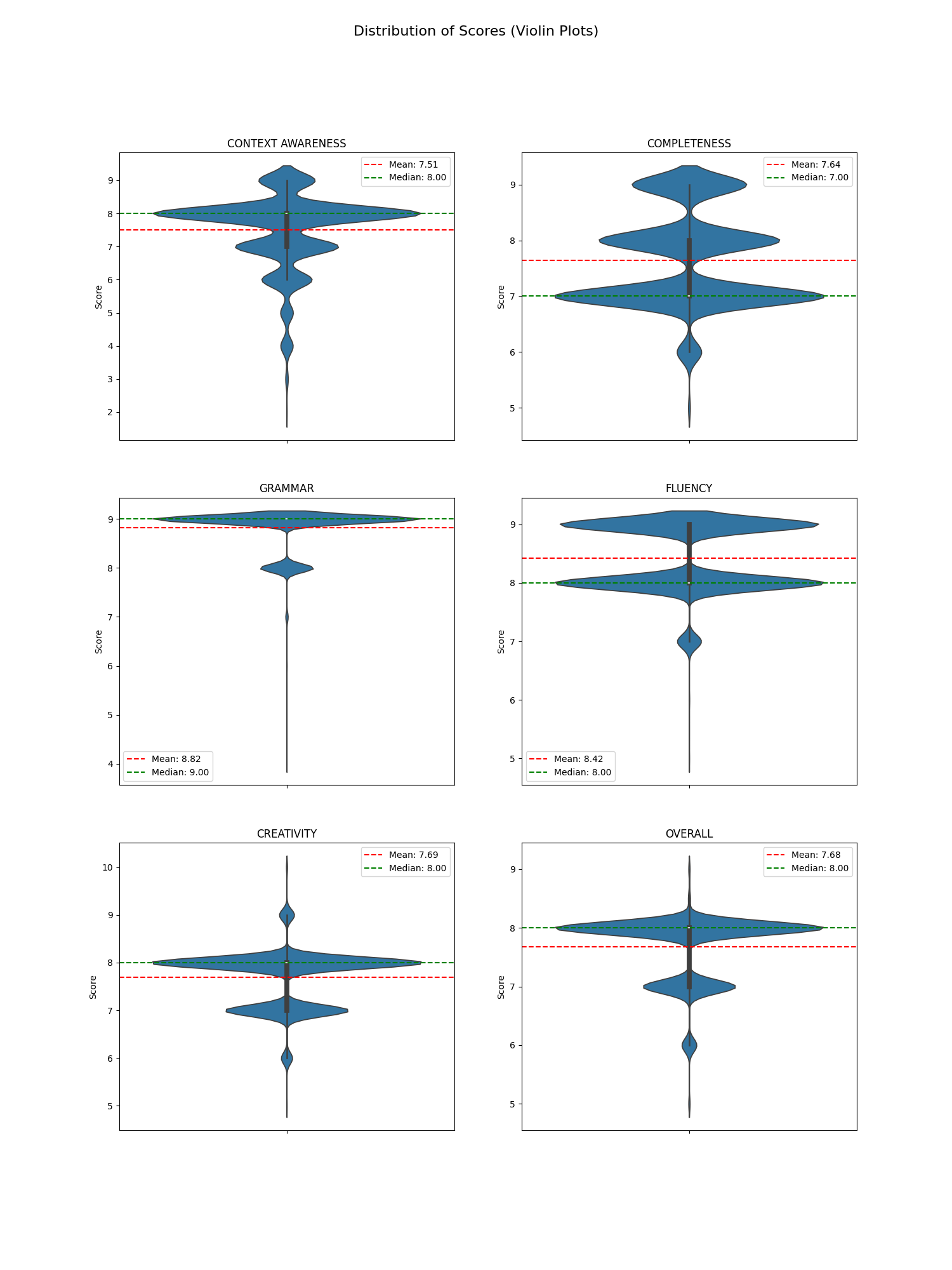}
        \caption{Bengali 54M inference score violin plots, n = 3000 }
        \label{fig:figure4}
    \end{minipage}
\end{figure}

\clearpage
\section{Generated story samples for Marathi and Bengali}

Figures 10-13 show the stories generated from
our TinyStories 54M model and that of GPT-4o for Marathi and Bengali language for the same prompt. The prompt which we provided effectively translates to:

\begin{tcolorbox}[colback=gray!20, colframe=gray!50, boxrule=0.5pt, arc=2mm]
‘‘Once upon a time, there was a small boy. His toy...’’
\end{tcolorbox}

\begin{figure}[ht]
    \centering
    \begin{minipage}{0.48\textwidth} 
        \centering
        \includegraphics[width=\linewidth]{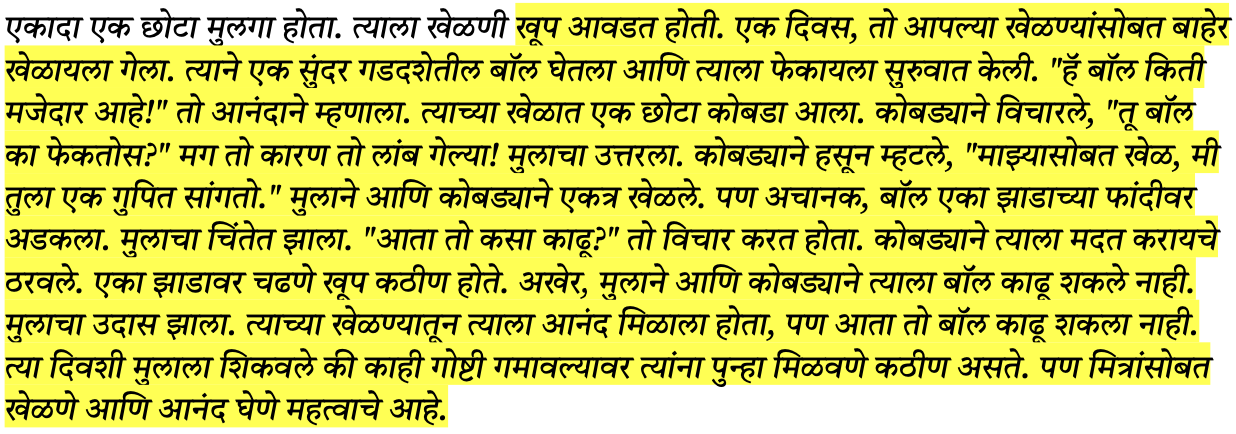}
        \caption{Regional TinyStories 54M Marathi model generated story}
        \label{fig:comparison_marathi_54M}
    \end{minipage}
    \hfill
    \begin{minipage}{0.48\textwidth}
        \centering
        \includegraphics[width=\linewidth]{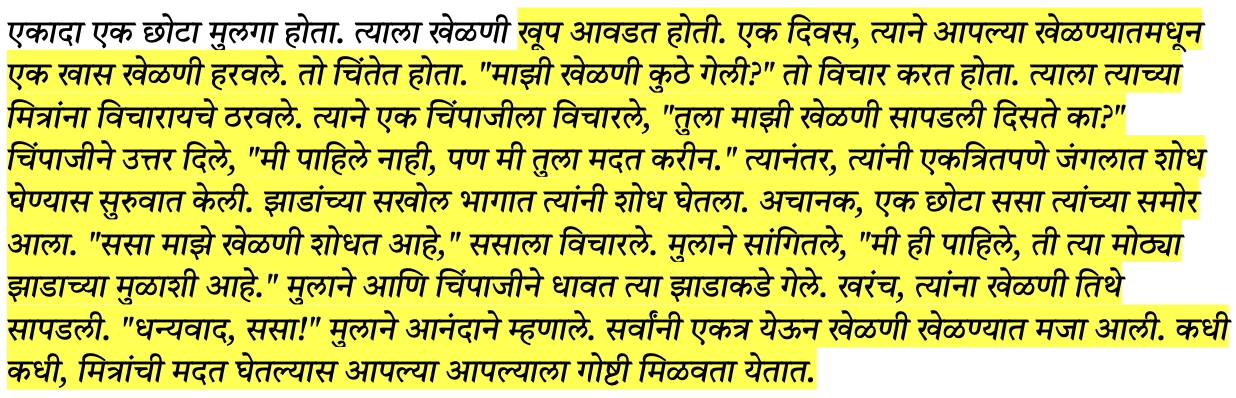}
        \caption{GPT 4o generated Marathi story}
        \label{fig:comparison_marathi_gpt4o}
    \end{minipage}
\end{figure}

\begin{figure}[ht]
    \centering
    \begin{minipage}{0.48\textwidth}
        \centering
        \includegraphics[width=\linewidth]{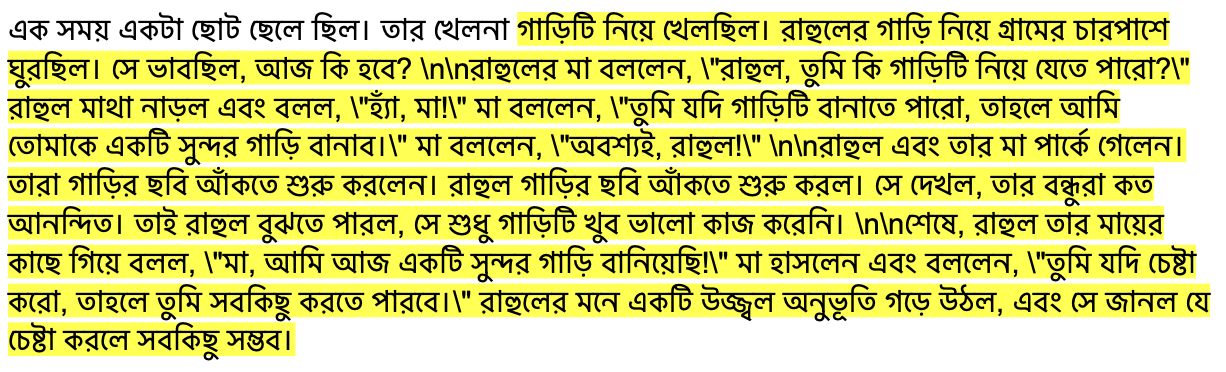}
        \caption{Regional TinyStories 54M Bengali model generated story}
        \label{fig:comparison_bengali_5M}
    \end{minipage}
    \hfill
    \begin{minipage}{0.48\textwidth}
        \centering
        \includegraphics[width=\linewidth]{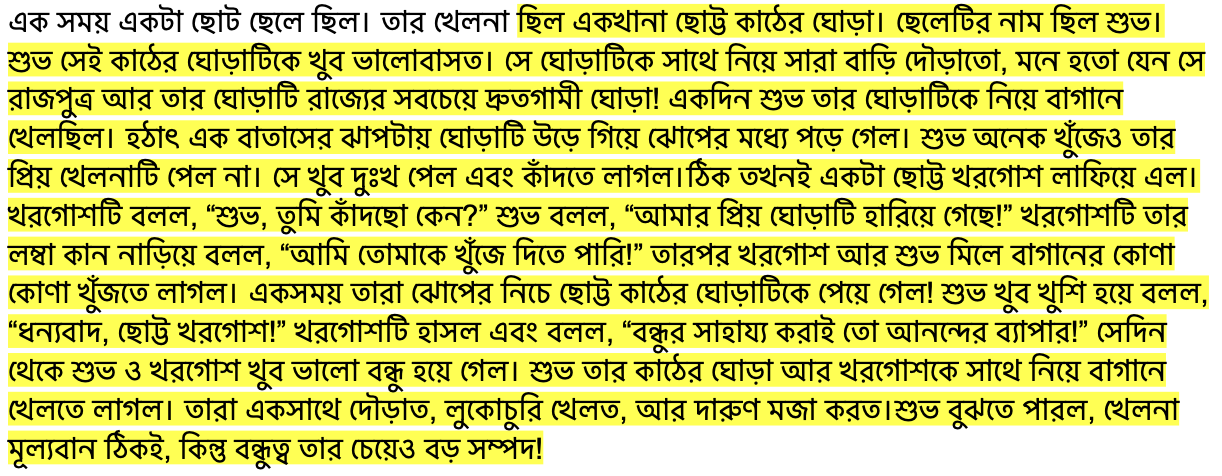}
        \caption{GPT 4o generated Bengali story}
        \label{fig:comparison_bengali_GPT4o}
    \end{minipage}
\end{figure}

We then ask GPT-4o to qualitatively compare between the 2 stories. Here is the output for Marathi language:

\begin{tcolorbox}[colback=gray!20, colframe=gray!50, boxrule=0.5pt, arc=2mm]
Overall, the Text A on the left offers more playful dialogues and scene details around a lost ball, while Text B on the right provides a succinct adventure‐style narrative about finding a missing pen. Both convey simple morals about seeking and giving help, but they differ in their degree of description, the nature of supporting characters, and the pacing of the action.
\end{tcolorbox}

Here is the output for Bengali language:

\begin{tcolorbox}[colback=gray!20, colframe=gray!50, boxrule=0.5pt, arc=2mm]
Overall, Text A on the left (story by 54M model) revolves around a boy named Rahul and his toy car, focusing on everyday interactions and the excitement he feels learning to help and explore. Text B on the right (story by GPT 4.0), on the other hand, highlights a boy’s wooden horse and leans into imaginative, almost fairy‐tale elements where the child envisions himself as a prince. Both narratives depict a child’s sense of wonder and creativity, yet Text A remains more grounded in everyday life, while Text B draws on a more whimsical, dream‐like tone to convey its central theme of playful discovery.
\end{tcolorbox}

\clearpage
\section{Synthetic dataset generation through LLM prompting}

\subsection{Dataset Generation Strategy}

The prompt generation process began with creating comprehensive lexical resources for each target language: Hindi, Bengali, and Marathi. We compiled vocabulary lists consisting of approximately 300 nouns, 300 verbs, and 300 adjectives appropriate for children aged 5-7 years for each of the languages. These were stored in language-specific text files.

Additionally, we developed "features" lists in both English and the target languages. These features represented narrative elements, themes, or tones to guide story generation (e.g., learning values, friendship themes, acts of kindness). These resources were consolidated into a structured JSON format for each language.

\subsection{Unique Prompt Generation Algorithm}

To ensure maximum diversity in the dataset while preventing duplicates, we implemented Algorithm 1.

\begin{algorithm}
\caption{Unique Prompt Generation}
\label{alg:prompt_generation}
\begin{flushleft}
\textbf{Input:} $N_{words}$ (nouns), $V_{words}$ (verbs), $A_{words}$ (adjectives), $F_{words}$ (features), $TargetCount$ (prompt count)\\
\textbf{Output:} $P$ (unique prompts)\\
\end{flushleft}
\begin{algorithmic}[1]
\STATE $UsedIDs \leftarrow \emptyset$
\STATE $UsedTriplets \leftarrow \emptyset$
\STATE $P \leftarrow \emptyset$
\STATE $DuplicateCount \leftarrow 0$
\WHILE{$|P| < TargetCount$}
    \STATE $n \leftarrow$ Select random element from $N_{words}$
    \STATE $v \leftarrow$ Select random element from $V_{words}$
    \STATE $a \leftarrow$ Select random element from $A_{words}$
    \STATE $f \leftarrow$ Select random element from $F_{words}$
    \STATE $ID \leftarrow ConcatenateIndices(n, v, a, f)$
    \STATE $TripletID \leftarrow ConcatenateIndices(n, v, a)$
    \IF{$ID \notin UsedIDs$ and $TripletID \notin UsedTriplets$}
        \STATE $UsedIDs \leftarrow UsedIDs \cup \{ID\}$
        \STATE $UsedTriplets \leftarrow UsedTriplets \cup \{TripletID\}$
        \STATE $prompt \leftarrow FormatTemplate(n, v, a, f)$
        \STATE $P \leftarrow P \cup \{prompt\}$
    \ELSE
        \STATE $DuplicateCount \leftarrow DuplicateCount + 1$
    \ENDIF
\ENDWHILE
\STATE \textbf{return} $P$, $DuplicateCount$
\end{algorithmic}
\end{algorithm}

This approach effectively prevented repetition patterns in the dataset, eliminating approximately 37,500 potential duplicate prompts from the target 3M dataset per language. The tracking of both quadruplet and triplet identifiers ensured maximum lexical diversity in the stories.

\subsection{Prompt Complexity Evolution}

We systematically explored different prompt complexity levels to identify the optimal configuration for generating high-quality children's stories. Five distinct complexity levels were developed, with increasing sophistication:

\begin{itemize}
    \item \textbf{Level 1}: Basic structure (TinyStories baseline) with minimal guidance
    \item \textbf{Level 2}: Enhanced structure with explicit narrative guidance (beginning/middle/end) and tone constraints
    \item \textbf{Level 2+}: Extended word limit (350-500 words) while maintaining structural guidance
    \item \textbf{Level 3}: Addition of dialogue elements (maximum three exchanges) and thematic guidance
    \item \textbf{Level 4}: Incorporation of cultural references (e.g., Panchatantra, Tenali Raman stories)
    \item \textbf{Level 4+/5}: Extension with supporting characters and natural elements
\end{itemize}
Through comparative evaluation using GPT-4 as the assessment model, complexity level 2+ was determined to provide the optimal balance of quality and generation efficiency. This template consistently yielded stories that achieved evaluation scores averaging 8.73 across all metrics (completeness, grammar, fluency, creativity).

\subsection{Optimal Prompt Template}

The Level 2+ template that produced the best results across languages followed this structure:

\begin{figure}[h]
\begin{center}
\fbox{
\begin{minipage}{0.95\textwidth}
\textbf{Optimal Prompt Template (Level 2+):}\\[0.5em]
Write a short story in \{language\} suitable for 5-to-7-year-old children.\\[0.3em]

Use simple, easy-to-understand words and limit the story to 3-4 short paragraphs (around 350-500 words).\\[0.3em]

The story should feature a clear beginning, middle, and end.\\[0.3em]

Incorporate the verb "\{verb\}", the noun "\{noun\}", and the adjective "\{adjective\}" naturally into the story.\\[0.3em]

The story should also integrate the conclusion/tone "\{feature\}" through actions and outcomes, without directly stating the tone.\\[0.3em]

Remember to use only simple words and keep the story appropriate for the target age group.\\[0.3em]

Return the output as a JSON dictionary: \{ "story": "your\_generated\_story" \}
\end{minipage}
}
\caption{Template used for story generation prompts across all three languages}
\label{fig:prompt_template}
\end{center}
\end{figure}

This template's effectiveness stems from several critical elements:

\begin{enumerate}
    \item It specifies a clear target audience and language
    \item It provides explicit structural guidance (3-4 paragraphs, clear beginning/middle/end)
    \item It incorporates lexical constraints (verb, noun, adjective) to guide vocabulary usage
    \item It requests thematic integration (feature/tone) through narrative rather than explicit statements
    \item It maintains appropriate word count constraints (350-500 words)
    \item It specifies the return format (JSON) for consistent processing
\end{enumerate}

\subsection{Implementation and Data Generation}

The prompt generation process was designed for scalability. For each language,  3 million unique prompts were generated and stored in JSON files. The implementation included progress tracking and efficient JSON-based storage.

For the actual story generation, a parallel processing approach with multiple concurrent API sessions was employed. We could configure 4 concurrent sessions, each with 16 threads, to achieve an approximate generation rate of 100 stories per minute with GPT-4o-mini as the generation model.

After evaluating multiple models (GPT-4o, LLaMA-3.1 70B, Claude 3.5 Sonnet),  GPT-4o-mini was selected (Fig. 15) based on its optimal balance of quality and generation efficiency, consistently achieving an 8.5/10 average score across evaluation metrics.

The final dataset included 2.2 million synthetic stories each for Hindi, Bengali, and Marathi, all generated using this systematic approach. This data generation methodology ensured both diversity and quality in the Regional TinyStories dataset, enabling effective training of Small Language Models for these languages.

\begin{figure}[ht]
    \centering
    \includegraphics[width=1.0\textwidth]{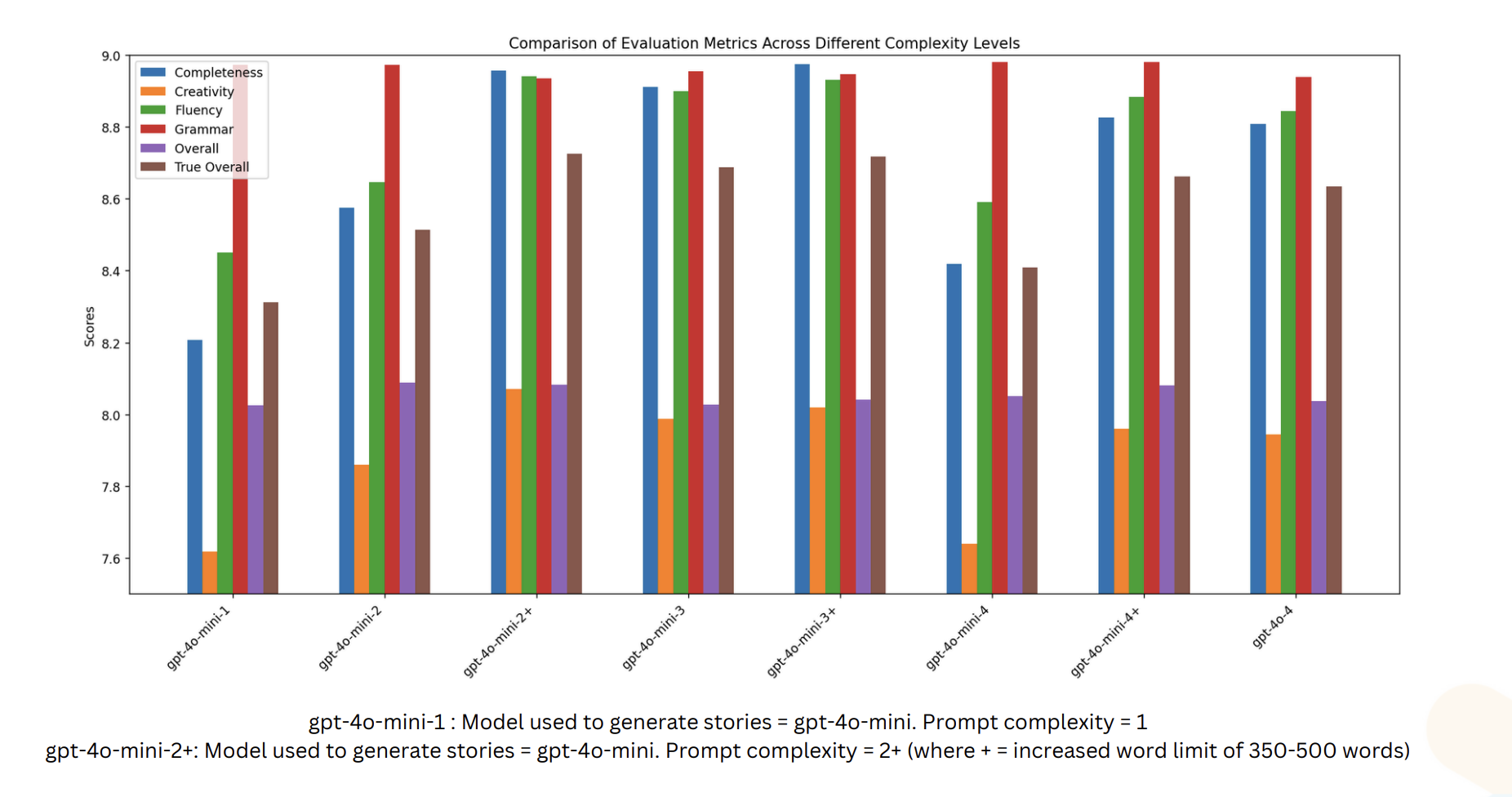}
    \caption{Comparison of Evaluation metrics across different complexity levels }
    \label{fig:prompt complexity evaluation metrics}
\end{figure}

\section{Training Data Analysis : Linguistic Diversity and Evaluation Metric Performance}

\subsection{The Zero-ROUGE Phenomenon in Cross-Lingual Evaluation}

Our experiments revealed a striking phenomenon when applying traditional n-gram based evaluation metric like ROUGE\cite{lin-2004-rouge} to non-English text generation. ROUGE (Recall-Oriented Understudy for Gisting Evaluation) is a set of metrics designed to evaluate automatic summarization and machine translation by comparing generated text to reference texts. ROUGE-1 and ROUGE-2 measure the overlap of unigrams (single words) and bigrams (word pairs) respectively between the candidate and reference texts, while ROUGE-L uses the longest common subsequence to assess sentence-level structural similarity. We wanted to utilize ROUGE to analyze the diversity / quality of the synthetically generated training dataset for our SLMs.
Although evaluating text generation quality for English has established benchmarks, we observed significant challenges when applying the same metrics to, e.g., LLM generated Bengali training stories from the TinyStories-Regional dataset.

\subsubsection{Contrasting ROUGE Performance Between Languages}

When applied to the English TinyStories dataset \cite{eldan2023tinystoriessmalllanguagemodels}, ROUGE metrics provided nuanced scores reflecting different degrees of lexical overlap:

\begin{verbatim}
Average ROUGE Scores (English):
Average ROUGE-1 F1: 0.2916
Average ROUGE-2 F1: 0.0553
Average ROUGE-L F1: 0.1700
\end{verbatim}

Individual story scores exhibited a normal distribution of values, matching the reports from the Tinystories paper:

\begin{table}[h]
\centering
\caption{English TinyStories ROUGE scores sample}
\label{tab:eng_rouge}
\begin{tabular}{|c|c|c|c|}
\hline
story\_idx & rouge1\_f1 & rouge2\_f1 & rougeL\_f1 \\
\hline
0 & 0.272727 & 0.054054 & 0.124579 \\
1 & 0.258503 & 0.006849 & 0.102041 \\
2 & 0.375000 & 0.094488 & 0.218750 \\
$\ldots$ & $\ldots$ & $\ldots$ & $\ldots$ \\
9 & 0.266160 & 0.061303 & 0.152091 \\
\hline
\end{tabular}
\end{table}

\sloppy 

However, when the same methodology was applied to the Bengali TinyStories dataset 
(\texttt{TinyStories-Regional/} \texttt{beng-generated\_4o-mini\_2M}), 
ROUGE uniformly produced zero values:

\begin{verbatim}

Average ROUGE Scores (Bengali):
Average ROUGE-1 F1: 0.0000
Average ROUGE-2 F1: 0.0000
Average ROUGE-L F1: 0.0000
\end{verbatim}

This striking result persisted across all ten pairs of evaluated stories, with precision, recall, and F1 scores uniformly at zero for all ROUGE variants.

\subsubsection{Contextual Analysis of Evaluation Metric Performance}

To better understand this phenomenon, we conducted a comprehensive comparative analysis using other evaluation metrics. BLEU (Bilingual Evaluation Understudy) \cite{10.3115/1073083.1073135} is an algorithm for evaluating machine translation quality based on n-gram precision, measuring how many generated phrases match reference translations. BLEU scores range from 0 to 1, with higher values indicating closer alignment to human references, though the metric tends to favor shorter texts and often fails to capture semantic equivalence.

BERTScore \cite{zhang2020bertscoreevaluatingtextgeneration} leverages contextual embeddings from pre-trained language models to compute similarity between generated and reference texts at a semantic level rather than exact word matches. This approach allows BERTScore to recognize paraphrases and synonyms as similar, making it more robust for evaluating text generation quality in morphologically rich languages where lexical variation is common.

METEOR (Metric for Evaluation of Translation with Explicit ORdering) \cite{banerjee-lavie-2005-meteor} evaluates translation quality by calculating precision and recall weighted by importance, while also accounting for word order, stemming, and synonymy. METEOR typically correlates better with human judgments than BLEU by considering linguistic elements beyond n-gram matching, making it particularly useful for evaluating text in languages with flexible word order and rich morphology.

\paragraph{BLEU Score Analysis.} BLEU scores for Bengali stories exhibited considerable variation yet remained consistently low. The mean BLEU score was 0.078 ($\sigma=0.126$), with values ranging from 0.003 to 0.421. Story index 5 demonstrated a notably higher BLEU score (0.421), suggesting some lexical alignment with its reference. The generally low BLEU scores corroborated our ROUGE findings, confirming significant lexical divergence (Fig. 15).

\paragraph{BERTScore Analysis.} In stark contrast to lexical metrics, BERTScore values were remarkably high across all Bengali story pairs. The mean BERTScore was 0.967 ($\sigma=0.012$), with scores ranging from 0.944 to 0.982. This dramatic difference between BLEU and BERTScore revealed a fundamental characteristic of the generated stories: while they utilize different vocabulary and phrasing from the references, they maintain high semantic fidelity (Fig. 15).

\paragraph{METEOR Score Analysis.} METEOR scores occupied a middle ground between BLEU and BERTScore, with a mean of 0.153 ($\sigma=0.046$) and range of 0.071 to 0.231.  For the sample in Fig. 16, story index 2 achieved the highest METEOR score (0.231), while story index 9 received the lowest (0.071). The intermediate nature of METEOR scores reflects its design as a balanced metric that considers both lexical and semantic similarities.

\begin{figure}[!ht]
    \centering
    \begin{minipage}[t]{\textwidth}
        \centering
        \includegraphics[width=0.7\textwidth]{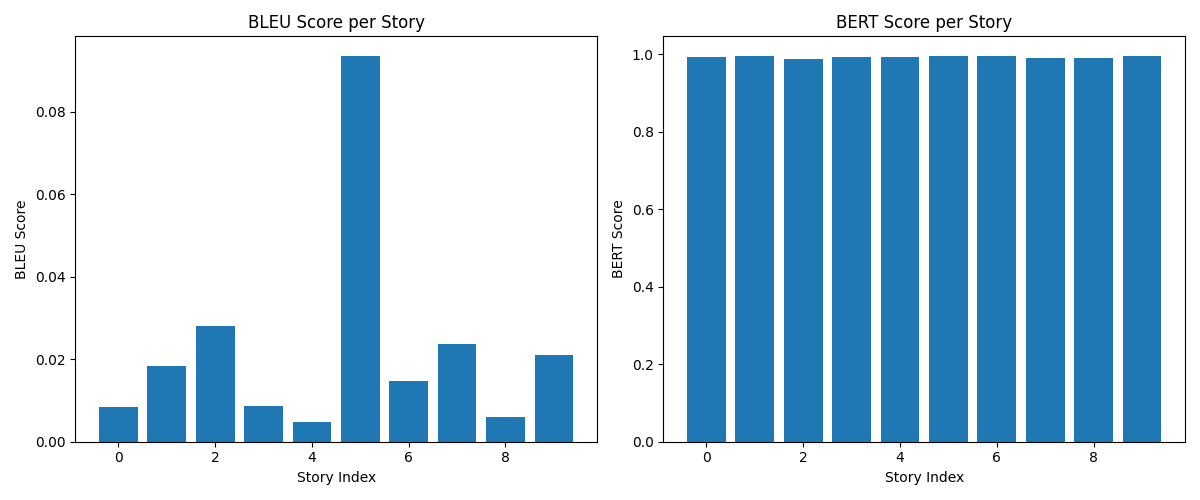}
        \caption{BLEU and BERT scores for 10 randomly selected stories from the synthetic Bengali dataset}
        \label{fig:figure3}
    \end{minipage}
    
    \vspace{1cm} 
    
    \begin{minipage}[t]{\textwidth}
        \centering
        \includegraphics[width=0.7\textwidth]{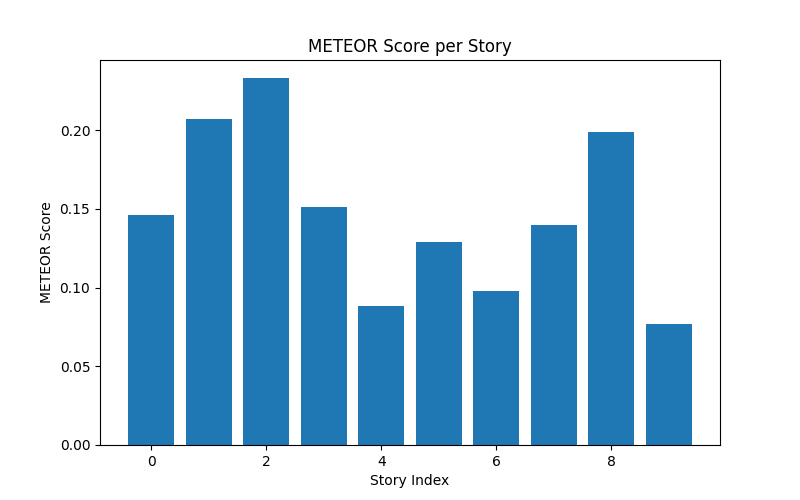}
        \caption{METEOR scores for 10 randomly selected stories from the synthetic Bengali dataset}
        \label{fig:figure4}
    \end{minipage}
\end{figure}

\FloatBarrier 
\newpage
\begin{figure}[!ht]
    \centering
    \begin{minipage}[t]{\textwidth}
        \centering
        \includegraphics[width=0.7\textwidth]{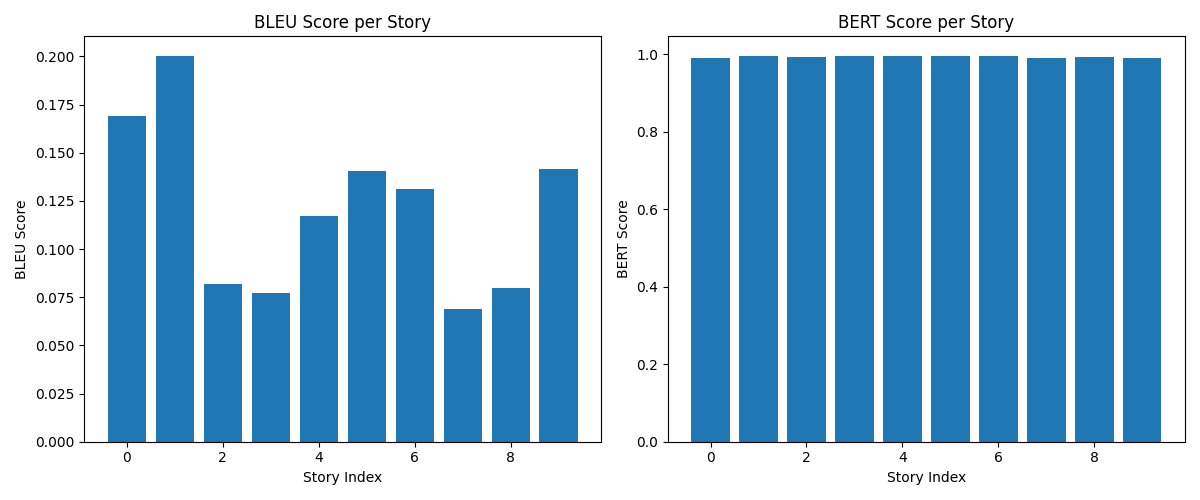}
        \caption{BLEU and BERT scores for 10 randomly selected stories from the synthetic Hindi dataset}
        \label{fig:figure3}
    \end{minipage}
    
    \vspace{1cm} 
    
    \begin{minipage}[t]{\textwidth}
        \centering
        \includegraphics[width=0.7\textwidth]{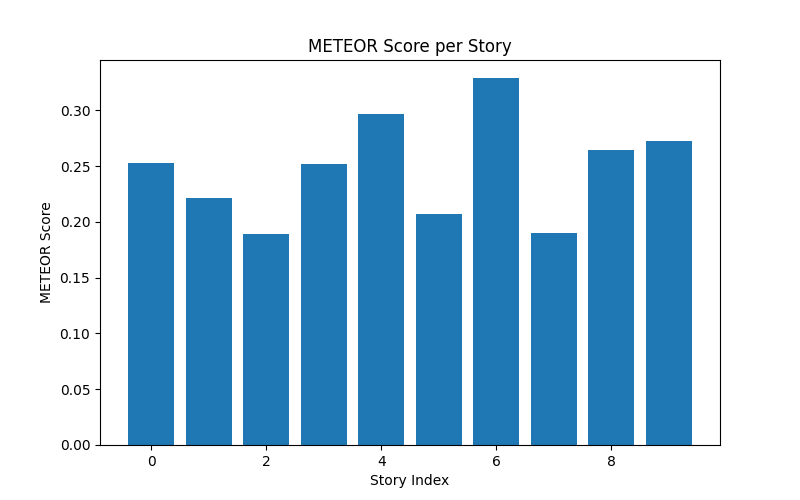}
        \caption{METEOR scores for 10 randomly selected stories from the synthetic Hindi dataset}
        \label{fig:figure4}
    \end{minipage}
\end{figure}

\FloatBarrier
\newpage

\begin{figure}[!ht]
    \centering
    \begin{minipage}[t]{\textwidth}
        \centering
        \includegraphics[width=0.7\textwidth]{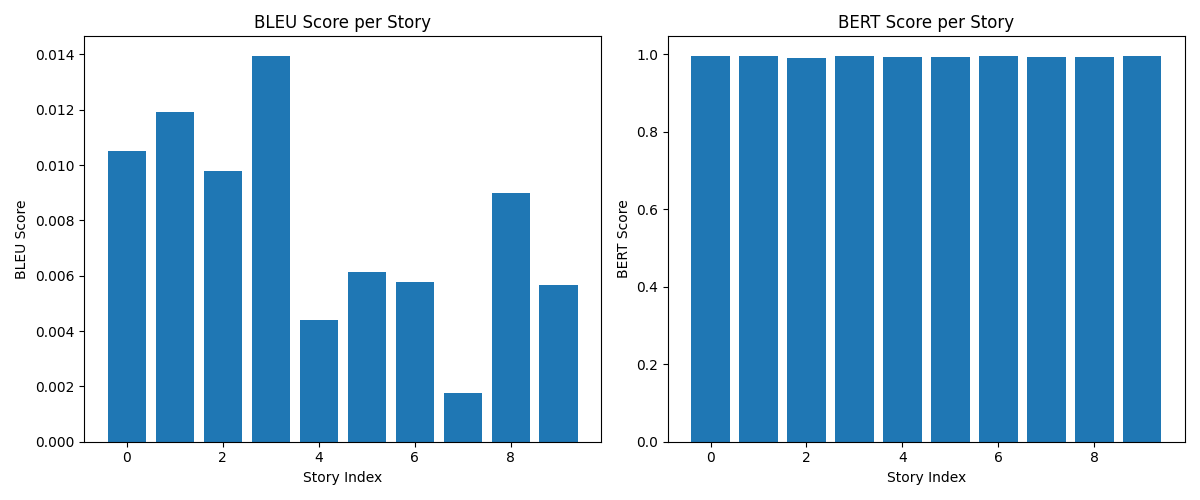}
        \caption{BLEU and BERT scores for 10 randomly selected stories from the synthetic Marathi dataset}
        \label{fig:figure3}
    \end{minipage}
    
    \vspace{1cm} 
    
    \begin{minipage}[t]{\textwidth}
        \centering
        \includegraphics[width=0.7\textwidth]{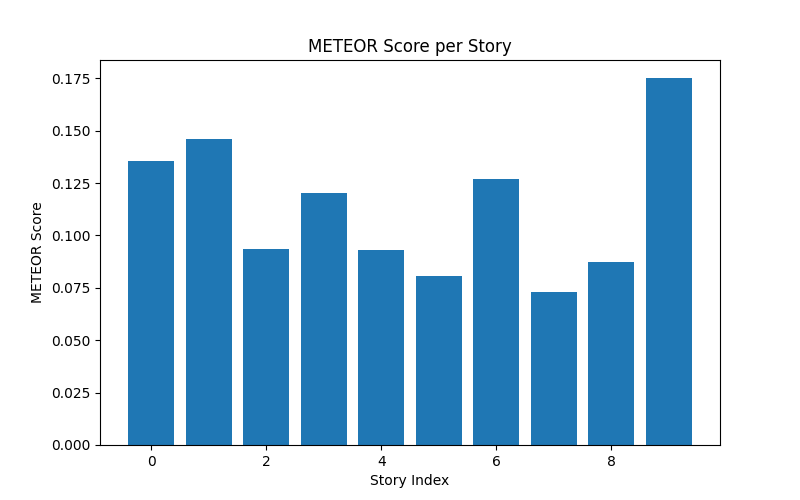}
        \caption{METEOR scores for 10 randomly selected stories from the synthetic Marathi dataset}
        \label{fig:figure4}
    \end{minipage}
\end{figure}

\FloatBarrier 

\newpage

\begin{table}[h]
\centering
\caption{Pearson Correlation Coefficients Between Metrics}
\label{tab:metric_corr}
\begin{tabular}{|l|c|}
\hline
Metric Pair & Correlation Coefficient \\
\hline
BLEU-BERTScore & 0.29 \\
BLEU-METEOR & 0.63 \\
BERTScore-METEOR & 0.51 \\
\hline
\end{tabular}
\end{table}

The moderate correlation between BLEU and METEOR ($r=0.63$) suggests that despite METEOR's consideration of synonymy, it still maintains sensitivity to lexical overlap. The weaker correlation between BLEU and BERTScore ($r=0.29$) confirms that these metrics capture fundamentally different aspects of text similarity. Qualitatively similar observations hold true for randomly sampled synthetic training data in Hindi and Marathi as observed in Figs. 17-20.

\subsection{Linguistic Factors Contributing to the Zero-ROUGE Phenomenon}

The zero-ROUGE phenomenon observed in Bengali text evaluation can be attributed to several linguistic factors:

\begin{enumerate}
\item \textbf{Morphological Richness:} Bengali possesses a complex morphological structure with numerous inflectional and derivational forms, increasing the likelihood of lexical variation even when expressing identical concepts.

\item \textbf{Word Formation Patterns:} The agglutinative tendencies in Bengali create fewer opportunities for exact n-gram matches compared to English.

\item \textbf{Syntactic Flexibility:} Bengali permits greater variation in word order while preserving meaning, reducing the likelihood of matching n-grams even in semantically equivalent sentences.

\item \textbf{Training Methodologies:} Modern language models with multiple decoding paths may naturally produce diverse lexical realizations of similar semantic content, especially when the target language permits such variation.
\end{enumerate}

This finding represents an extreme manifestation of the limitations of lexical metrics, where the absence of exact n-gram overlap, as evidenced by zero ROUGE scores, suggests that text generation systems employ sophisticated paraphrasing mechanisms while maintaining semantic coherence.

\subsection{Implications for Multi-Lingual Text Generation Evaluation}

Our analysis suggests that robust evaluation of text generation requires a multi-metric, language-aware approach. Based on our findings, we propose:

\subsubsection{Language-Specific Considerations}

\begin{enumerate}
\item \textbf{Metric Selection:} Researchers must carefully select evaluation metrics appropriate to the target language, considering morphological complexity and typical paraphrasing patterns.

\item \textbf{Benchmark Calibration:} Distinct performance benchmarks should be established for each language rather than applying universal thresholds derived from English.

\item \textbf{Reference Design:} Evaluation datasets for morphologically rich languages should include multiple reference texts to better capture acceptable lexical variation.
\end{enumerate}

\subsubsection{Multi-Dimensional Evaluation Framework}

For comprehensive assessment of generated text quality across languages, we recommend an integrated approach:

\begin{enumerate}
\item \textbf{Semantic Fidelity Assessment:} Using embedding-based metrics like BERTScore with language-specific models to verify preservation of core meaning.

\item \textbf{Structural Evaluation:} Employing METEOR with language-appropriate resources for stemming and synonymy to assess whether narrative structure and word order are maintained within language-specific constraints.

\item \textbf{Lexical Diversity Measurement:} Calculating type-token ratios or using metrics like MTLD \citep{McCarthy2010} to quantify lexical richness relative to language norms.

\item \textbf{Reference-Free Quality Assessment:} Incorporating fluency and coherence metrics calibrated to the specific language being evaluated.
\end{enumerate}

\subsection{Case Study: Qualitative Analysis of Bengali Story Pairs}

To illustrate the disconnect between lexical overlap and semantic similarity, we present a representative Bengali story pair from our dataset, alongwith the English translations in Figs. 21 and 22:

\begin{figure}[ht]
    \centering
    \includegraphics[width=1.0\textwidth]{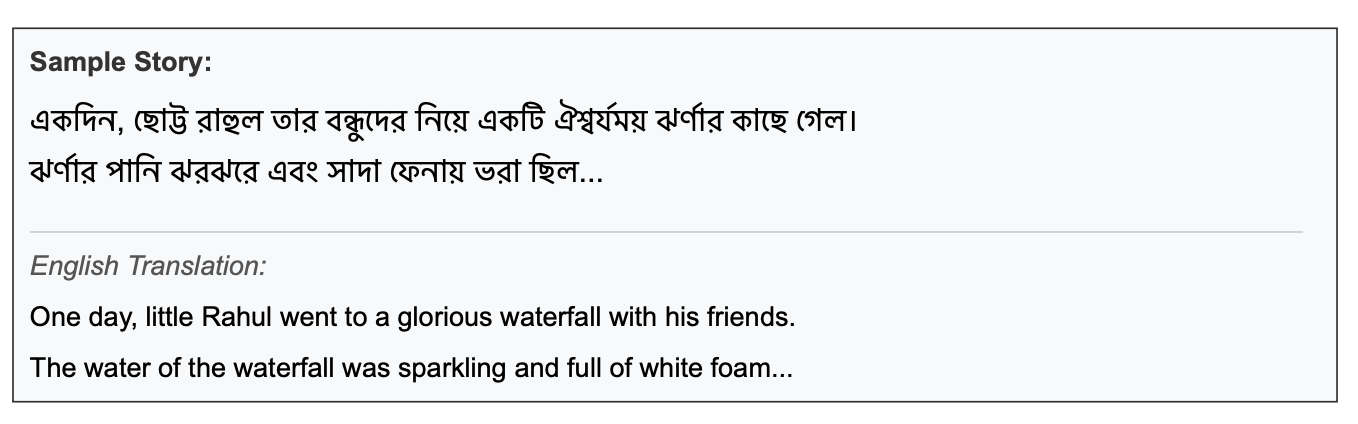}
    \caption{A sample Bengali story from the synthetic dataset}
    \label{fig:sample Bengali story}
\end{figure}

\begin{figure}[ht]
    \centering
    \includegraphics[width=1.0\textwidth]{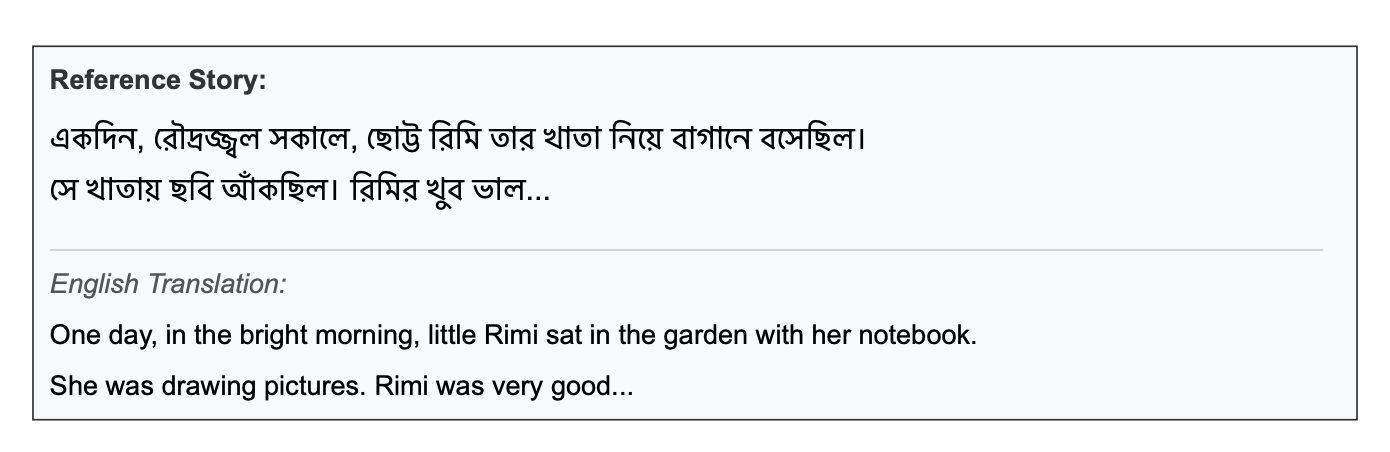}
    \caption{A story for reference from the same dataset }
    \label{fig:ref Beng story}
\end{figure}
Despite sharing the theme of a child's experience outdoors, these stories use entirely different vocabulary, characters, and settings. ROUGE metrics registered zero overlap, yet BERTScore identified high semantic similarity (0.961), recognizing the shared narrative elements and emotional tone.

\subsection{Final comments}

Our discovery of the zero-ROUGE phenomenon highlights the need for better evaluation frameworks for non-English languages, particularly as text generation systems prioritize semantic preservation over lexical copying.

Analysis across Hindi, Bengali, and Marathi reveals consistent patterns:
\begin{itemize}
    \item BLEU scores remain low ($<$0.2)
    \item BERTScore values approach near-perfect ($>$0.95)
    \item METEOR scores provide a middle ground (0.07-0.33)
\end{itemize}
This contrast demonstrates how traditional lexical metrics fail to capture semantic equivalence in morphologically rich Indian languages. The pattern confirms our generation approach produces semantically coherent content with lexical diversity, rather than relying on exact phrase repetition.

Future research directions should include:

\begin{itemize}
    \item Developing specialized metrics balancing plot preservation with stylistic variation
    \item Establishing multilingual benchmark datasets with multiple reference texts
    \item Investigating human-metric correlations for generative tasks
    \item Exploring reference-free evaluation approaches
\end{itemize}

\section{Details on translated dataset and inference evaluations}

\subsection{Translation method}

The original TinyStories dataset in English \cite{eldan2023tinystoriessmalllanguagemodels} has a train split that contains 2,119,719 stories, with 320,470 duplicates, reducing unique stories to 1,799,249 (15.12\% duplicates). This may impact model training, potentially skewing results. 

Examples of stories appearing three times include :

\begin{itemize}
    \item "Lily and Ben were playing in the park. They saw a fox hiding in the bushes. The fox had red fur and ..."
    \item "Tom is a boy who likes to play with his lab. His lab is a big, black dog who can do amazing tricks. ..."
    \item "Sara loved her pet cat, Lily. Lily was soft and fluffy and liked to sleep on Sara's bed. Sara liked ..."
    \item "Sam was sick. He had a bad cough and a sore throat. He did not like being sick. He wanted to play wi..."
\end{itemize}

Validation split has 21,990 unique stories without duplicates, ensuring reliable evaluation. However, cross-split analysis shows 6,601 stories appear in both splits, which could inflate model performance metrics. Addressing these duplicates is crucial for accurate model comparison between translated and synthetic data, enhancing the robustness of our findings in regional language modeling. Hence, we removed the duplicates and merged the two datasets. Additionally, due to resource constraints we could only translate for Hindi and Bengali. Apart from the previously mentioned issue with duplicates, we observed some data loss in translation due to issues with the freely available Google Translate API, resulting in approximately 1.9 M unique stories in Hindi and Bengali. Due to severe resource constraint, we could not translate to Marathi.

It was decided to use GPT4o for rating the translation quality based on recent reports \cite{kocmi2023largelanguagemodelsstateoftheart, jiao2023chatgptgoodtranslatoryes} of GPT4 matching or outperforming Google Translate in case of non-English languages. In the context of evaluating translation models for Hindi and Bengali, using GPT4o, several multilingual models were assessed to identify the most effective translation method. The models include mBART \cite{liu2020multilingual}, IndicTrans2 \cite{gala2023indictrans2}, mT5 \cite{xue2020mt5}, Helsinki-NLP OPUS MT \cite{tiedemann2023democratizing}, IndicBART \cite{dabre2021indicbart}, NLLB \cite{nllbteam2022languageleftbehindscaling}, and M2M100 \cite{fan2021beyond}. Each model offers distinct capabilities, particularly in handling translations between English and Indic languages. GPT4o was given original English text and asked to translate to the target Indic language. Following this, it was provided with one of the machine translations and asked to rate it out of 10, using the following prompt :

\begin{figure}[H]
    \centering
    \begin{tcolorbox}[colback=gray!15, colframe=gray!25, boxrule=0.5pt, arc=2mm, width=\linewidth]
       I have used a machine translation model to translate the original story. On a scale of 1-10 evaluate the translation quality of the story, with respect to your translation, 1 being very bad and 10 being of the same quality as yours. Remember that "quality" here does not mean the same words but meaningfully retaining the same context and fluency. Also, point out each instance where there is a mistake in translation.
    \end{tcolorbox}
\end{figure}

For Bengali translation, IndicTrans2  achieved an average score of 7/10 on 100 stories. Google Translate, although widely used, sometimes produced translations with contextual inaccuracies and verb errors. Still it was rated at an average score of 8 by GPT-4o. Hence it was chosen for Bengali. On the other hand, the NLLB model received an average score of 7/10. GPT-4o noted that while the NLLB translations were generally good, they lacked the fluency and natural phrasing found in Google Translate. Specific issues were identified with verb tenses and word choices, leading to minor awkwardness in sentence structures. 

We used a Python library called DeepTranslate to perform Google Translate API calls. This ran on CPU hardware but we performed several optimizations to improve translation speed, namely batch processing of stories, rate limiting at 5 requests per second, and robust fault tolerance measures. These measures included automatic retries with exponential backoff, an LRU cache for successful translations, and regular intermediate saves. Quality control involved clear story demarcation, UTF-8 encoding, and progress tracking. 

However, for Hindi translations both NLLB and Google Translate produced scores around 8.5. Hence, the total Hindi transaltion was split between these two, primarily due to resource planning regarding API calls for Google Translate and GPU time for NLLB.

Overall, these evaluations demonstrate the strengths and weaknesses of different models in translating between English and Indic languages, highlighting the importance of choosing a model that balances semantic accuracy with natural language fluency. This analysis is critical for understanding the challenges and opportunities in model translation quality, particularly in academic contexts where precise language use is essential. 

\subsection{Inference scores}
\begin{table*}[hbt]
\renewcommand{\thetable}{14}
\label{tab:translated vs synthetic}
\scriptsize
\centering
\begin{tabular}{c c c c c c c c}
\toprule
\textbf{Dataset used}& \textbf{Eval Loss} & \textbf{Context} & \textbf{Completeness} & \textbf{Creativity} & \textbf{Fluency} & \textbf{Grammar} & \textbf{Overall} \\
\midrule
\multicolumn{8}{l}{\textbf{Hindi}} \\
\midrule
Translated data& \ColorEvalLossCell{1.3849}{red}& \ColorGeneralCell{6.178}{red}& \ColorGeneralCell{5.806}{red}& \ColorGeneralCell{5.938}{red}& \ColorGeneralCell{6.597}{red}& \ColorGeneralCell{7.438}{red}& \ColorGeneralCell{6.391}{red}\\
Synthetic data& \ColorEvalLossCell{0.518}{red} & \ColorGeneralCell{7.734}{red} & \ColorGeneralCell{7.783}{red} & \ColorGeneralCell{7.806}{red} & \ColorGeneralCell{8.554}{red} & \ColorGeneralCell{8.912}{red} & \ColorGeneralCell{8.158}{red} \\
\midrule
\multicolumn{8}{l}{\textbf{Bengali}} \\
\midrule
Translated data
& \ColorEvalLossCell{1.4939}{beige}& \ColorGeneralCell{6.879}{beige}& \ColorGeneralCell{6.598}{beige}& \ColorGeneralCell{6.462}{beige}& \ColorGeneralCell{7.339}{beige}& \ColorGeneralCell{8.122}{beige}& \ColorGeneralCell{7.080}{beige}\\
Synthetic data& \ColorEvalLossCell{0.569}{beige} & \ColorGeneralCell{7.507}{beige} & \ColorGeneralCell{7.645}{beige} & \ColorGeneralCell{7.693}{beige} & \ColorGeneralCell{8.420}{beige} & \ColorGeneralCell{8.816}{beige} & \ColorGeneralCell{8.016}{beige} \\
\end{tabular}
\caption{\small{Inference score evaluations for Hindi and Bengali in case of models trained with translated vs synthetic dataset. Model configuration : 6 layers, 8 attention heads, 512 hidden embeddings. 3000 stories were evaluated for each model. Its observed that models trained with translated data have reduced performance.}}
\label{tab:translated vs synthetic}
\end{table*}
\citeauthor{boughorbel2024improvinglanguagemodelstrained} \citeyearpar{boughorbel2024improvinglanguagemodelstrained} explored how machine-translated data can be used to train language models for generating stories in Arabic, and pinpointed problems related to linguistic and cultural biases present in the translated data. They suggested additional pre-training of the models using a limited amount of high-quality synthetic data and examined the impact of this approach using Dictionary Learning techniques. They further reported that the resulting Sparse Auto-Encoders demonstrated a change in the features they learn, showing improved linguistic characteristics and less cultural bias. In any event, we have shown that training on fully synthetic data that has been generated with high quality vetting process can indeed result in significant performance improvements as shared in Table 14. The drop in quality of inference for Hindi compared to Bengali could be due to two separate translation means employed for generating Hindi translated data. We do not have the resource at present to thoroughly evaluate this. 

\newpage
\section{WeightWatcher Analysis of Bengali Short Story Generation Model}

\subsection{Overview}
This appendix presents a quantitative analysis of the Bengali short story generation model using WeightWatcher (WW) \cite{Martin2021}, a tool designed to assess the quality and stability of neural network weights. WW analyzes statistical properties of weight matrices to identify potential issues with training and generalization. It analyzes the quality of deep learning model layers by computing layer-specific metrics, including the $\alpha$ metric. This metric is based on the Heavy-Tailed Spectral Random (HTSR) theory, which suggests that well-trained layers exhibit a specific spectral density shape. In the context of WeightWatcher, a good $\alpha$ value for a well-trained layer is generally considered to be between 2 and 6. 

\subsection{Model Architecture Summary}
The analyzed model is a transformer-based architecture with approximately 157M parameters, consisting of 7 transformer blocks, each with attention and feed-forward components. The model uses a 1024-dimensional embedding space and features dense linear projections throughout its architecture. In terms of overall score, this was the best performing Bengali model (Table 3). The distribution of layer $\alpha$ values are shown in Fig. 23 and  Table 14 provides a summary of the model architecture post-training. 

\begin{figure}[ht]
    \centering
    \includegraphics[width=0.6\textwidth]{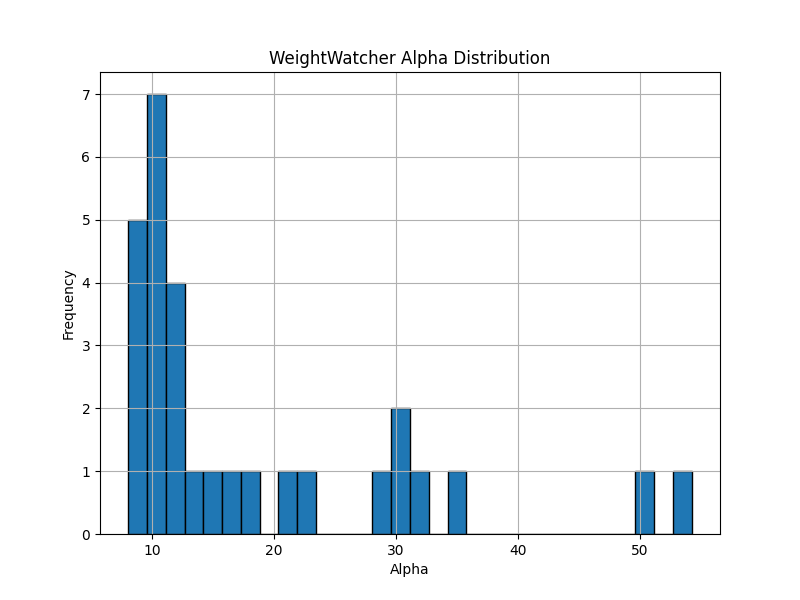}
    \caption{Distribution of $\alpha$ values of layers of the 157M parameter Bengali model }
    \label{fig:Beng model WW analysis}
\end{figure}

\begin{table}[h]
\centering
\caption{Model Architecture Summary}
\begin{tabular}{lr}
\hline
\textbf{Metric} & \textbf{Value} \\
\hline
Under-trained Layers & 29/29 (100\%) \\
Mean $\alpha$ & 22.51 \\
$\alpha$ Range & 6.76 - 89.34 \\
Mean Entropy & 0.9655 \\
Mean Spectral Norm & 1.7654 \\
Mean Stable Rank & 392.25 \\
\hline
\end{tabular}
\end{table}

\subsection{Layer Type Analysis}
The model consists of one embedding layer and 28 dense layers. The dense layers include attention query, key, and value projections, attention output projections, and feed-forward network components. Table \ref{tab:layer_types} presents metrics across different layer types.

\begin{table}[h]
\centering
\caption{Metrics by Layer Type}
\label{tab:layer_types}
\begin{tabular}{lrrrr}
\hline
\textbf{Layer Type} & \textbf{Count} & \textbf{Mean $\alpha$} & \textbf{Mean Spectral Norm} & \textbf{Mean Entropy} \\
\hline
Embedding & 1 & 6.76 & 1.6231 & 0.9280 \\
Dense & 28 & 23.07 & 1.7705 & 0.9669 \\
\hline
\end{tabular}
\end{table}

\subsection{Notable Observations}

\subsubsection{Alpha Values ($\alpha$)}
The Power-Law (PL) exponent $\alpha$ is a key metric in WeightWatcher's analysis, with higher values potentially indicating instability or over-parameterization. Our model shows generally high $\alpha$ values (mean: 22.51), with several extreme outliers in deeper layers of the network. Three layers exhibit particularly high values:

\begin{table}[h]
\centering
\caption{Layers with Outlier Alpha Values}
\label{tab:alpha_outliers}
\begin{tabular}{lrr}
\hline
\textbf{Layer} & \textbf{$\alpha$} & \textbf{Spectral Norm} \\
\hline
transformer.h.2.mlp.c\_fc & 70.94 & 3.6742 \\
transformer.h.4.mlp.c\_proj & 89.34 & 0.2627 \\
transformer.h.6.mlp.c\_proj & 79.97 & 0.2652 \\
\hline
\end{tabular}
\end{table}

These extreme values, particularly in later transformer blocks, suggest potential instability in the model's deeper layers. Notably, the projection layers in MLP blocks exhibit the highest $\alpha$ values, indicating they may be problematic components in the training process.

\subsubsection{Spectral Norms}
The distribution of spectral norms demonstrates a clear pattern where attention query and MLP feed-forward layers have consistently higher spectral norms (mean $\approx$ 3.0-3.7), while attention projection and MLP projection layers have much lower values (mean $\approx$ 0.1-0.3). This dichotomy reflects the architectural design of transformer networks where projection layers typically compress information.

\subsubsection{Entropy Values}
Entropy values are consistently high across all layers (mean: 0.9655), suggesting good information propagation through the network. Dense layers show slightly higher entropy values compared to the embedding layer, which is expected given their role in transforming and processing the information flow through the model.

\subsection{Analysis of Transformer Blocks}
Examining metrics across the 7 transformer blocks reveals interesting patterns:

\begin{enumerate}
\item Blocks 2, 4, and 6 show notably higher mean $\alpha$ values, suggesting potential instability in these specific blocks.
\item Spectral norms remain relatively consistent across blocks, indicating a stable architectural design throughout the model depth.
\item The number of PL spikes (indicating heavy-tailed eigenvalue distributions) decreases in deeper layers, which may indicate diminishing expressivity in later layers.
\end{enumerate}

\subsection{Recommendations for Model Improvement}
Based on the WeightWatcher analysis, we recommend the following steps to potentially enhance model performance:

\begin{enumerate}
\item \textbf{Extended Training:} All layers show signs of under-training, suggesting the model could benefit from additional training epochs.
\item \textbf{Layer-specific Learning Rates:} Apply differential learning rates for problematic layers with extremely high $\alpha$ values, particularly the MLP projection layers in transformer blocks 2, 4, and 6.
\item \textbf{Regularization Strategies:} Consider layer-specific regularization techniques for components with outlier metrics to stabilize their behavior.
\item \textbf{Architecture Refinement:} Potential benefit from architectural modifications to blocks with extreme $\alpha$ values, such as adjusting the hidden dimensions or introducing additional normalization.
\end{enumerate}

\subsection{Final comments}
The WeightWatcher analysis provides valuable insights into the Bengali short story generation model's internal characteristics. The model demonstrates high entropy values (mean: 0.9655), indicating effective information propagation, although several layers in transformer blocks 2, 4, and 6 exhibit extremely high alpha values (70.94-89.34), which may indicate instability in these components. We observed a clear dichotomy in spectral norms between attention query and MLP feed-forward layers (3.0-3.7) versus projection layers (0.1-0.3), reflecting the architectural design of information compression in transformer networks. Based on these findings, we recommend targeted regularization strategies for outlier layers and potentially adjusting learning rates for problematic MLP projection components to enhance model stability and generative performance.


\end{document}